\documentclass{article}

\usepackage{microtype}
\usepackage{graphicx}
\usepackage{booktabs}
\usepackage{multirow}
\usepackage{xcolor}
\usepackage{amsmath,amssymb,mathtools}
\usepackage{amsthm}
\usepackage{times}  
\usepackage{newtxmath}  
\usepackage{bm}
\usepackage{float}  
\usepackage{hyperref}

\newif\ificmlstyle
\IfFileExists{icml2026.sty}{\icmlstyletrue}{\icmlstylefalse}  

\newif\ifsubmission
\submissionfalse  

\ificmlstyle
  \ifsubmission
    \usepackage{icml2026}            
  \else
    \usepackage[preprint]{icml2026}  
  \fi
  \hypersetup{pdfsubject={}}
\else
  \usepackage[letterpaper,margin=0.75in,columnsep=0.25in]{geometry}
  \usepackage{times}
  \usepackage{natbib}
  \newcommand{\icmltitlerunning}[1]{}
  \newcommand{\icmltitle}[1]{{\centering\Large\bfseries #1\par}\vskip 0.2in}
  \newenvironment{icmlauthorlist}{\centering}{\par}
  \newcommand{\icmlauthor}[2]{#1\quad}
  \newcommand{\icmlaffiliation}[2]{}
  \newcommand{\icmlcorrespondingauthor}[2]{}
  \newcommand{\icmlkeywords}[1]{}
  \newcommand{\printAffiliationsAndNotice}[1]{}
\fi

\usepackage[capitalize,noabbrev]{cleveref}
\theoremstyle{plain}
\newtheorem{theorem}{Theorem}[section]
\newtheorem{proposition}[theorem]{Proposition}
\newtheorem{corollary}[theorem]{Corollary}

\definecolor{gain}{RGB}{0,120,60}
\newcommand{\pmstd}[2]{#1\,{\scriptsize$\pm$#2}}
\newcommand{\up}[2]{\textcolor{gain}{+#1}}
\DeclareMathOperator*{\Median}{median}
\DeclareMathOperator{\sg}{sg}
\newcommand{\norm}[1]{\left\lVert #1 \right\rVert}
\newcommand{\hz}{\hat{z}}
\newcommand{\Cz}{\mathcal{C}_0}
\newcommand{\Ctot}{\mathcal{C}_{\mathrm{EKR}}}
\newcommand{\M}{\mathcal{M}}
\newcommand{\cRef}{c_{\mathrm{ref}}}
\newcommand{\EKR}{EKR}

\icmltitlerunning{Escaping Latent Traps by Remembering}

\begin{document}

\twocolumn[
\icmltitle{Escaping Latent Traps by Remembering: Training-Free\\
           Episodic Kernel Repulsion for World-Model Planning}

\begin{icmlauthorlist}
\icmlauthor{Yican Zhao}{qias,zju}
\icmlauthor{Bolin Yang}{qias,zju}
\icmlauthor{Ruizhe Wang}{qias}
\icmlauthor{Haiyang Guo}{qias,zju} \\
\icmlauthor{Fuji Yang}{qias}
\icmlauthor{Bingo Wing-Kuen Ling}{qias,gdut}
\icmlauthor{Zelin Zang}{qias}
\end{icmlauthorlist}
\icmlaffiliation{qias}{Institute of Integrated Circuits and Artificial Intelligence, Tsientang Institute for Advanced Study, Hangzhou, China}
\icmlaffiliation{zju}{Zhejiang University, Hangzhou, China}
\icmlaffiliation{gdut}{Guangdong University of Technology, Guangzhou, China}
\icmlcorrespondingauthor{Zelin Zang}{zangzelin@gmail.com}
\icmlkeywords{world models, model predictive control, JEPA, planning, exploration}

\vskip 0.3in
]
\printAffiliationsAndNotice{}

\begin{abstract}
Latent world models plan towards a goal image by minimising the distance
between predicted and goal latents. Because this distance is not geodesic, it
creates \emph{latent traps}---local minima or flat basins behind walls, at
corners and in dead ends---where memoryless receding-horizon replanning keeps
revisiting states far from the goal.
We propose \emph{Episodic Kernel Repulsion} (\EKR{}), a training-free,
test-time cost term that penalises candidate plans via a top-$k$ Gaussian
kernel to a novelty-gated memory of the latents the agent has \emph{actually
visited}, normalised by the current batch base cost so a single strength
transfers across models.
Across two JEPA world models and four benchmarks, \EKR{} improves success in
all eight settings without changing the model, planner or search budget,
matching or exceeding $4\times$ samples at $0.8$--$1.5\times$ compute in
seven of eight settings while adding ${\sim}25$\,ms ($<1\%$) per replan.
Controlled trap-fork interventions confirm that the memory causes escape, and
an untuned configuration transfers directly to DINO-WM with CEM (e.g.\
$44.7\%\to 100.0\%$ on long PointMaze, $+19.3$ points on long Wall) and to
JEPA-WM on Metaworld \texttt{mw-reach-wall} ($25.3\%\to 40.7\%$).
\end{abstract}

\section{Introduction}
\label{sec:intro}

World models trained by latent prediction~\citep{lecun2022path,assran2023ijepa},
such as DINO-WM, LeWM and H-JEPA~\citep{zhou2025dinowm,maes2026lewm,zhang2026hjepa},
have become a practical route to zero-shot visual control.
\begin{figure}[t]
\centering
\includegraphics[width=\columnwidth]{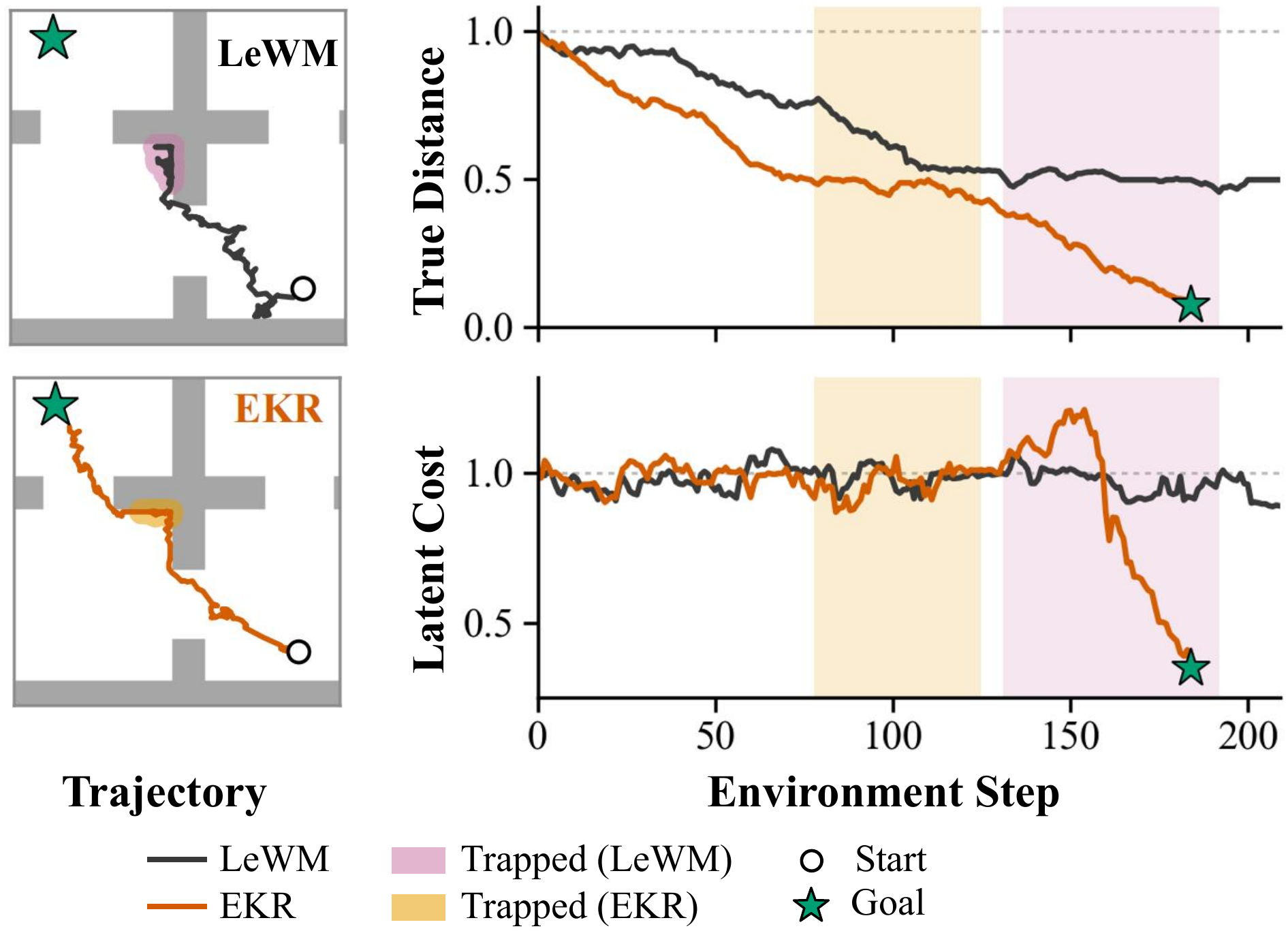}
\caption{\textbf{Escaping a latent trap on AntMaze.}
Left: top-down trajectories of the baseline (LeWM, grey) and \EKR{} (orange)
from the start (circle) to the goal (star); shaded patches mark the trapped
region of each run. Right: true distance to the goal and latent cost, both
normalised by their initial value, against the environment step; the shaded
bands mark the trapped steps of \EKR{} (yellow) and of the baseline (pink).
The baseline stalls with the true distance near half its initial value, while
\EKR{} leaves the trap and reaches the goal at about step 183. Its latent cost
first rises, then falls once the goal is in reach.}
\label{fig:trap}
\end{figure}
An encoder maps the current observation and a goal image to latents, and a
model-predictive controller (MPC) searches for actions whose predicted latent
trajectory ends close to the goal latent. Nothing beyond the world model is
learned: no reward, no value function, no policy.

The price is that the planner sees the task only through this latent
distance to the goal, the \emph{base cost}. With obstacles, this distance is
not geodesic and creates \emph{latent traps}---local minima or flat basins
behind a wall, at a corner or in a dead end---where every short-horizon
rollout that leaves the region looks worse than staying. Receding-horizon replanning does not escape them,
because the cost depends only on the current state and the goal, not on what
the agent has already tried; the executed plan returns the agent to states
from which the same plans are chosen again (\cref{sec:formulation}).
\Cref{fig:trap} shows a typical case: on AntMaze the baseline enters a side
pocket and stalls at about half its initial distance to the goal.

Existing remedies change how the planner searches, not what it knows. A
larger budget finds lower-cost plans, but inside a trap the lowest-cost plans
are those that stay: for any history-independent planner, extra budget only
changes the per-step escape probability, which does not grow however long the
agent has been trapped (Prop.~\ref{prop:memoryless}). Collocation, hybrid
optimisers, graph search and hierarchical
subgoals~\citep{rybkin2021latco,bharadhwaj2020cemgd,eysenbach2019sorb,zhang2026hjepa}
modify the optimiser or add a learned component, and hierarchical models
still get trapped (\cref{sec:main}). What the planner lacks is a record of
where the agent has already been.

We propose \emph{Episodic Kernel Repulsion} (\EKR{}), which turns this record
into a planning signal. \EKR{} stores the latents of states the agent has
actually visited in a novelty-gated episodic memory and adds to the base cost
a term that repels candidate plans from them. Plans that lead back into a
visited basin become more expensive with every revisit, so the basin fills up
until an exit becomes the cheapest option. The repulsion is a top-$k$
Gaussian kernel, so a growing memory does not dilute it, and it is scaled by
the current base cost, so a single strength $\lambda$ applies across models
whose costs differ by orders of magnitude. \EKR{} needs no training, no extra
network and no change to the planner or its budget.

\paragraph{Contributions.}
\begin{itemize}
  \setlength\itemsep{1pt}
  \item \textbf{Problem.} We identify latent traps as a failure mode of
        latent-world-model MPC, caused by a non-geodesic cost under
        memoryless replanning, and show that for any history-independent
        planner the escape probability does not grow with the time spent in
        a trap (\cref{sec:formulation}, \cref{app:theory}).
  \item \textbf{Method.} We propose \EKR{}, a training-free cost term that
        repels plans from the agent's visited states, and prove that it
        bounds the time spent in traps of bounded depth
        (\cref{sec:method}, Thm.~\ref{thm:sojourn}).
  \item \textbf{Evidence.} Across LeWM and H-JEPA on four benchmarks,
        \EKR{} improves success in all eight settings under the official
        protocol and, in seven, matches or exceeds $4\times$ search at
        ${\le}1.5\times$ compute with $<1\%$ wall-clock overhead
        (\cref{sec:main,sec:compute,app:metaworld}). Causal trap-fork
        interventions, ablations and latent recurrence analyses show that
        memory of observed states drives escape
        (\cref{sec:ablation,sec:behavior,app:latent,app:diagnostics}).
  \item \textbf{Transfer and robustness.} Without tuning, the unified
        configuration transfers to DINO-WM with CEM, HWM hierarchies, and
        JEPA-WM on Metaworld \texttt{mw-reach-wall}, and holds across a broad
        hyperparameter range (\cref{sec:general,app:sensitivity,app:metaworld}).
\end{itemize}

\section{Related Work}
\label{sec:related}

\paragraph{Planning with latent world models.}
Latent dynamics models for planning go back to
PlaNet~\citep{hafner2019planet} and Dreamer~\citep{hafner2020dreamer};
TD-MPC~\citep{hansen2022tdmpc,hansen2024tdmpc2} adds a learned value and
reward. A newer line plans towards a goal image with a pure latent distance
cost: DINO-WM~\citep{zhou2025dinowm} on frozen
DINOv2~\citep{oquab2024dinov2} features with CEM~\citep{deboer2005cem},
V-JEPA~2-AC~\citep{assran2025vjepa2} for robot arms,
LeWM~\citep{maes2026lewm} end-to-end from pixels, and the hierarchical
HWM~\citep{zhang2026hwm} and H-JEPA~\citep{zhang2026hjepa}. All of them
replan with a cost that ignores the agent's history. \EKR{} leaves model,
hierarchy and planner untouched and only adds a history term to this cost.

\paragraph{Local minima in planning.}
Local minima of goal-directed costs are a classical failure of potential
fields~\citep{khatib1986real,koren1991potential}. In learned-model planning
they motivated collocation~\citep{rybkin2021latco}, hybrid CEM/gradient
planners~\citep{bharadhwaj2020cemgd}, graph search over a replay
buffer~\citep{eysenbach2019sorb} and hierarchical
subgoals~\citep{zhang2026hwm,zhang2026hjepa}, which change the optimiser or
add a learned component or level. \EKR{} is complementary: it changes only
the cost, using information the agent has already collected.

\paragraph{Exploration bonuses.}
Count-based bonuses~\citep{bellemare2016unifying,tang2017exploration},
RND~\citep{burda2019rnd}, episodic curiosity~\citep{savinov2019episodic} and
NGU~\citep{badia2020ngu} reward novel states to train exploratory policies;
Plan2Explore~\citep{sekar2020plan2explore} plans towards states where an
ensemble of world models disagrees, and Go-Explore~\citep{ecoffet2021first}
returns to promising states across episodes. Closest to \EKR{} is the
episodic term of NGU, also a $k$-nearest-neighbour kernel over an in-episode
memory. NGU, however, is an intrinsic reward that requires learned embeddings
and policy training, and it rewards states already reached. \EKR{} is
training-free: it scores the predicted end states of MPC candidates under a
frozen world model, and its scale is tied to the base cost. We compare with
count-based novelty directly in \cref{sec:ablation}.

\paragraph{Repulsion and memory in optimisation.}
Tabu search~\citep{glover1989tabu} forbids returning to recently visited
solutions of an optimiser, and novelty search~\citep{lehman2011abandoning}
replaces the objective by novelty relative to an archive. Repulsive particle
methods, such as SVGD~\citep{liu2016svgd}, Stein variational
MPC~\citep{lambert2020stein} and repulsive
ensembles~\citep{dangelo2021repulsive}, push candidates away from
\emph{each other} to keep them diverse. \EKR{} instead repels candidates from
the states the agent has \emph{executed} in the current episode, as a soft
cost added to an unchanged goal objective.

\section{Method}
\label{sec:method}

\begin{figure*}[t]
\centering
\includegraphics[width=\textwidth]{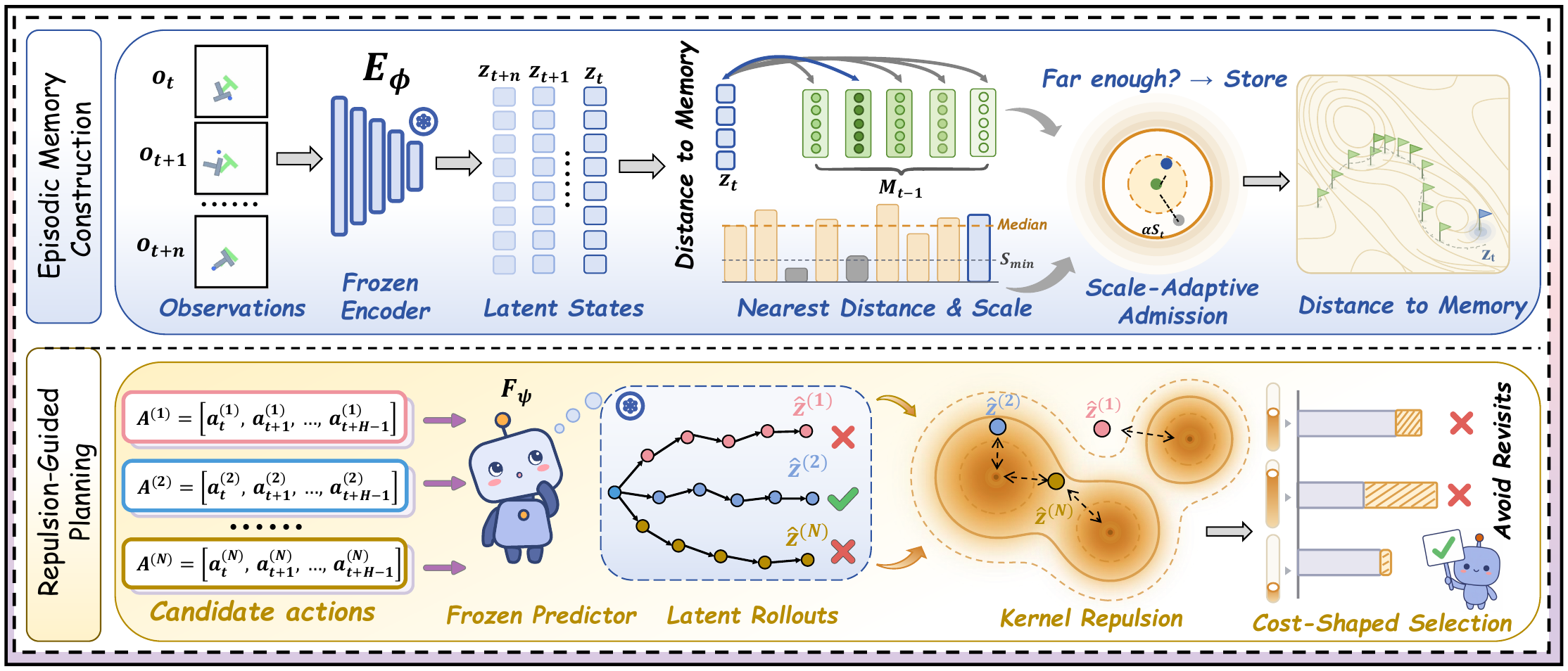}
\caption{\textbf{Overview of \EKR{}.} Top: episodic memory construction.
The frozen encoder $E_\phi$ maps observations to latent states; the distance
from the current latent $z_t$ to the memory $\mathcal{M}_{t-1}$ is compared
with a scale estimated from past nearest distances, and $z_t$ is stored only
if it is far enough. Bottom: repulsion-guided planning. The frozen predictor
$F_\psi$ rolls out $N$ candidate action sequences, a kernel repulsion term
penalises rollouts that end near stored states, and the cost-shaped selection
avoids revisits.}
\label{fig:method}
\end{figure*}

\subsection{Problem formulation}
\label{sec:formulation}
We consider goal-conditioned visual control with a frozen pretrained world
model. Its encoder $E_\phi$ maps the current and goal observations to latents
$z_t=E_\phi(o_t)$ and $z_g=E_\phi(o_g)$ in $\mathcal{Z}\subset\mathbb{R}^d$.
A model-predictive controller (MPC) optimises $N$ candidate action sequences
$\mathbf{a}^{(i)}=a^{(i)}_{t:t+H-1}$ over a horizon $H$, which the predictor
$F_\psi$ rolls out from $z_t$:
\begin{equation}
  \hz^{(i)}_{t+h}=F_\psi\big(\hz^{(i)}_{t+h-1},\,a^{(i)}_{t+h-1}\big),\qquad h=1,\dots,H.
  \label{eq:rollout}
\end{equation}
The base cost sums the squared distances to the goal latent over a cost
window $\mathcal{T}\subseteq\{1,\dots,H\}$,
\begin{equation}
  \Cz\big(\mathbf{a}^{(i)}\big)=\sum_{h\in\mathcal{T}}\norm{\hz^{(i)}_{t+h}-z_g}_2^2 .
  \label{eq:c0}
\end{equation}
The first action(s) of the best candidate are executed and the agent replans.

\begin{figure*}[t]
\centering
\includegraphics{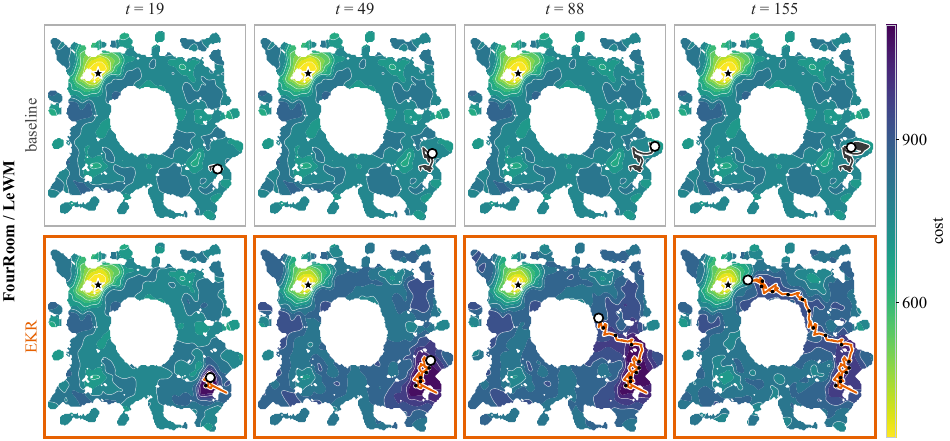}
\caption{\textbf{\EKR{} lifts the cost of visited states and breaks a latent
trap.} Representative FourRoom/LeWM task (seed 42, all doors open; baseline
fails, \EKR{} succeeds at $t{=}166$) at four steps ($M_t\in\{2,5,9,16\}$).
Costs are computed in the full latent space and shown on a 2-D Isomap layout
(display only; white: no data; shared scale $350$--$1110$). Top: static goal
cost $\Cz$; the baseline stalls at the lower right (white circle). Bottom:
\EKR{} adds memory repulsion (above its per-panel minimum, leaving $\arg\min$
unchanged) around stored states (black dots), lifting the visited region while
preserving the goal valley (star).}
\label{fig:landscape}
\end{figure*}

With obstacles, latent Euclidean distance differs from geodesic distance, so
$\Cz$ has local minima away from the goal. Because $\Cz$ depends only on
$(z_t,z_g)$, planning and executing induce a history-independent closed-loop
map $z_{t+1}=\Phi_0(z_t)$. We call a region $\mathcal{U}\subset\mathcal{Z}$
with
\begin{equation}
  \Phi_0(\mathcal{U})\subseteq\mathcal{U},\qquad z_g\notin\mathcal{U},
  \label{eq:trap}
\end{equation}
a \emph{latent trap} (\cref{fig:landscape}, top): once inside, the agent replans similar actions,
cycles, and never reaches the goal. A larger planning budget searches the
same cost landscape more finely; it can help in shallow traps but leaves
$\Phi_0$, and hence the invariant set, unchanged. Episodic Kernel Repulsion
(\EKR{}) is a training-free, test-time cost shaping that makes the planning
cost depend on the visitation history of the current episode, so that the
closed-loop map is no longer fixed and the invariance in \cref{eq:trap} can
be broken (\cref{fig:method}; formal analysis in \cref{app:theory}).

\subsection{Novelty-gated episodic memory}
\label{sec:memory}

\EKR{} keeps an episodic memory $\M_t=\{m_1,\dots,m_{M_t}\}\subset\mathcal{Z}$,
emptied at the start of every episode. It stores only latents of \emph{real}
observations, never predicted ones, so rollout errors cannot enter it; at
most one entry is written per replanning step, and beyond the capacity
$M_{\max}$ the oldest entry is dropped. The distance of the current latent to
the memory is
\begin{equation}
  d_{\min}(z_t,\M_{t-1})=\min_{m_j\in\M_{t-1}}\norm{z_t-m_j}_2 ,
  \label{eq:dmin}
\end{equation}
with $d_{\min}(\cdot,\emptyset)=\infty$. The spacing of the real trajectory
in latent space is estimated from recent nearest distances,
\begin{equation}
\begin{aligned}
  s_t&=\max\big(s_{\min},\ \beta\Median(\mathcal{D}_t)\big),\\
  \mathcal{D}_t&=\textstyle\bigcup_{b=1}^{B}\big\{d_{\min}\big(z^{b}_{\tau},\M^{b}_{\tau-1}\big)\big\}_{\tau=t-W+1}^{t},
\end{aligned}
  \label{eq:scale}
\end{equation}
where $W$ is the window length, $\beta>0$ a scale factor, $s_{\min}>0$ a
floor, and $b$ indexes the $B$ episodes run in parallel, which pool only this
statistic, not their memories. The novelty gate admits $z_t$ only if it is far
from the memory relative to this spacing,
\begin{equation}
  \M_t=
  \begin{cases}
    \M_{t-1}\cup\{z_t\}, & d_{\min}(z_t,\M_{t-1})>\alpha\, s_t ,\\
    \M_{t-1}, & \text{otherwise},
  \end{cases}
  \label{eq:admission}
\end{equation}
with $\alpha\in(0,1)$. Because the threshold scales with the spacing, local
jitter adds no entries, and the memory density reflects the coverage of the
visited region rather than the time spent in it.

\subsection{Local kernel repulsion potential}
\label{sec:potential}
\begin{table*}[t]
\centering
\caption{\textbf{Success rate (\%) under the official protocol.}
Mean\,$\pm$\,std over 3 seeds $\times$ 50 tasks. Held-out: 50 new tasks per
seed from the official distribution, configurations frozen. $\Delta$: absolute
gain; exact McNemar test on 150 paired episodes: $^{*}p{<}0.05$,
$^{**}p{<}0.01$, $^{***}p{<}0.001$.}
\label{tab:main}
\setlength{\tabcolsep}{7pt}
\renewcommand{\arraystretch}{1.05}
\begin{tabular}{cllcccc}
\toprule
 & Backbone & Method & AntMaze & FourRoom & Push-T & Cube \\
\midrule
\multirow{6}{*}{\rotatebox[origin=c]{90}{\emph{Official tasks}}}
 & \multirow{3}{*}{LeWM}
 & Baseline    & \pmstd{19.3}{6.1} & \pmstd{41.3}{14.7} & \pmstd{40.0}{3.5} & \pmstd{26.7}{1.2} \\
 & & + \EKR{} (ours) & \textbf{\pmstd{30.0}{9.2}}$^{*}$ & \textbf{\pmstd{62.7}{5.8}}$^{***}$ & \textbf{\pmstd{54.7}{4.6}}$^{**}$ & \textbf{\pmstd{39.3}{7.6}}$^{**}$ \\
 & & $\Delta$    & \up{10.7}{0} & \up{21.3}{0} & \up{14.7}{0} & \up{12.7}{0} \\
\cmidrule(l){2-7}
 & \multirow{3}{*}{H-JEPA}
 & Baseline    & \pmstd{66.0}{4.0} & \pmstd{92.7}{4.2} & \pmstd{10.0}{3.5} & \pmstd{50.0}{9.2} \\
 & & + \EKR{} (ours) & \textbf{\pmstd{73.3}{9.5}} & \textbf{\pmstd{97.3}{2.3}} & \textbf{\pmstd{21.3}{5.0}}$^{**}$ & \textbf{\pmstd{64.0}{12.0}}$^{*}$ \\
 & & $\Delta$    & \up{7.3}{0} & \up{4.7}{0} & \up{11.3}{0} & \up{14.0}{0} \\
\midrule
\multirow{6}{*}{\rotatebox[origin=c]{90}{\emph{Held-out tasks}}}
 & \multirow{3}{*}{LeWM}
 & Baseline    & \pmstd{25.3}{4.2} & \pmstd{34.7}{4.2} & \pmstd{34.7}{8.1} & \pmstd{19.3}{7.0} \\
 & & + \EKR{} (ours) & \textbf{\pmstd{38.7}{3.1}}$^{**}$ & \textbf{\pmstd{56.7}{4.2}}$^{***}$ & \textbf{\pmstd{46.0}{3.5}}$^{*}$ & \textbf{\pmstd{34.7}{4.6}}$^{**}$ \\
 & & $\Delta$    & \up{13.3}{0} & \up{22.0}{0} & \up{11.3}{0} & \up{15.3}{0} \\
\cmidrule(l){2-7}
 & \multirow{3}{*}{H-JEPA}
 & Baseline    & \pmstd{70.0}{8.7} & \pmstd{95.3}{3.1} & \pmstd{10.0}{3.5} & \pmstd{41.3}{4.2} \\
 & & + \EKR{} (ours) & \textbf{\pmstd{72.0}{12.5}} & \textbf{\pmstd{99.3}{1.2}} & \textbf{\pmstd{14.7}{8.3}} & \textbf{\pmstd{56.0}{6.0}}$^{*}$ \\
 & & $\Delta$    & \up{2.0}{0} & \up{4.0}{0} & \up{4.7}{0} & \up{14.7}{0} \\
\bottomrule
\end{tabular}
\end{table*}

Let $\hz^{(i)}=\hz^{(i)}_{t+H}$ be the predicted end latent of candidate $i$.
Its similarity to a memory entry is a Gaussian kernel with bandwidth $\sigma$
(\cref{sec:fusion}),
\begin{equation}
  \kappa_\sigma\big(\hz^{(i)},m_j\big)=\exp\!\Big(-\frac{\norm{\hz^{(i)}-m_j}_2^2}{2\sigma^2}\Big).
  \label{eq:kernel}
\end{equation}
Averaging the kernel over the whole memory would dilute the revisit signal
as the memory grows, because distant entries still enter the normaliser.
\EKR{} therefore uses only the $K_{\mathrm{eff}}=\min(K,M_t)$ entries
$m_{(1)},\dots,m_{(K_{\mathrm{eff}})}$ nearest to $\hz^{(i)}$:
\begin{equation}
  \rho_K\big(\hz^{(i)}\big)=\frac{1}{K_{\mathrm{eff}}}\sum_{k=1}^{K_{\mathrm{eff}}}\kappa_\sigma\big(\hz^{(i)},m_{(k)}\big)\in[0,1],
  \label{eq:rhok}
\end{equation}
with $\rho_K\equiv0$ for an empty memory, so the repulsion does not depend on
how many distant entries the memory holds (empirically, this choice matters
little; \cref{sec:ablation}). Where the neighbour set is unique and $\sigma$
is fixed, $\rho_K$ is differentiable with
\begin{equation}
\begin{aligned}
  \nabla_{\hz}\rho_K&=-\frac{1}{K_{\mathrm{eff}}\sigma^2}\sum_{k=1}^{K_{\mathrm{eff}}}\kappa_\sigma\big(\hz^{(i)},m_{(k)}\big)\big(\hz^{(i)}-m_{(k)}\big),\\
  \nabla_{\mathbf{a}}\rho_K&=J_{\mathbf{a}}^{\top}\nabla_{\hz}\rho_K,\qquad J_{\mathbf{a}}=\partial\hz^{(i)}/\partial\mathbf{a}.
\end{aligned}
  \label{eq:grad}
\end{equation}
The descent direction $-\nabla_{\hz}\rho_K$ pushes the candidate away from its
neighbouring memories, with $\norm{\nabla_{\hz}\rho_K}_2\le1/(\sigma\sqrt{e})$;
each term peaks at distance $\sigma$ and decays at a Gaussian rate, so distant
candidates are hardly repelled. Gradient-based planners backpropagate this
gradient through the frozen world model; sampling-based planners use $\rho_K$
to rank candidates.

\subsection{Adaptive bandwidth and cost fusion}
\label{sec:fusion}
\paragraph{Adaptive bandwidth.} $\sigma$ sets the radius of repulsion and is
set online in one of two ways. \emph{Endpoint calibration} sets, at every
cost evaluation,
\begin{equation}
  \sigma=\max\Big(\sigma_{\min},\ \tau_\sigma\Median_{1\le i\le N}\,\min_{m_j\in\M_t}\norm{\hz^{(i)}-m_j}_2\Big).
  \label{eq:bandwidth}
\end{equation}
When the floor is inactive, a candidate at the median nearest-memory distance
has kernel value $e^{-1/(2\tau_\sigma^2)}$ regardless of the latent scale, so
the kernel neither saturates nor vanishes. \emph{Observation calibration}
sets $\sigma=s_t$, matching the trajectory spacing. In both cases $\sigma$ is
constant within a cost evaluation and not differentiated.

\paragraph{Scale-free cost fusion.} The magnitude of $\Cz$ varies with the
backbone and the latent dimension, whereas $\rho_K\in[0,1]$ is dimensionless.
The reference scale is therefore the mean base cost of all candidates in the
batch, with gradients stopped:
\begin{equation}
  \cRef=\sg\Big(\frac{1}{BN}\sum_{b=1}^{B}\sum_{i=1}^{N}\Cz\big(\mathbf{a}^{(b,i)}\big)\Big).
  \label{eq:cref}
\end{equation}
The shaped planning cost is
\begin{equation}
  \Ctot\big(\mathbf{a}^{(i)}\big)=\Cz\big(\mathbf{a}^{(i)}\big)+\lambda\,\cRef\,\rho_K\big(\hz^{(i)}\big).
  \label{eq:total}
\end{equation}
Since $\rho_K$ is independent of the scale of $\Cz$, for any $c>0$
\begin{equation}
  \Cz\mapsto c\,\Cz\ \Longrightarrow\ \cRef\mapsto c\,\cRef,\quad \Ctot\mapsto c\,\Ctot ,
  \label{eq:invariance}
\end{equation}
so the candidate ranking is invariant to the scale of the base cost, and
$\lambda$ bounds the repulsion relative to the mean candidate cost.
\Cref{fig:landscape} illustrates this effect on FourRoom/LeWM: even when
$\Cz$ does not form a deep local pit ($748$--$779$ at the stall site),
repulsion lifts that region by $158$--$289$ as memory grows while leaving the
goal valley intact, steering the plan out of the trap.

\begin{figure*}[t]
\centering
\includegraphics[width=\textwidth]{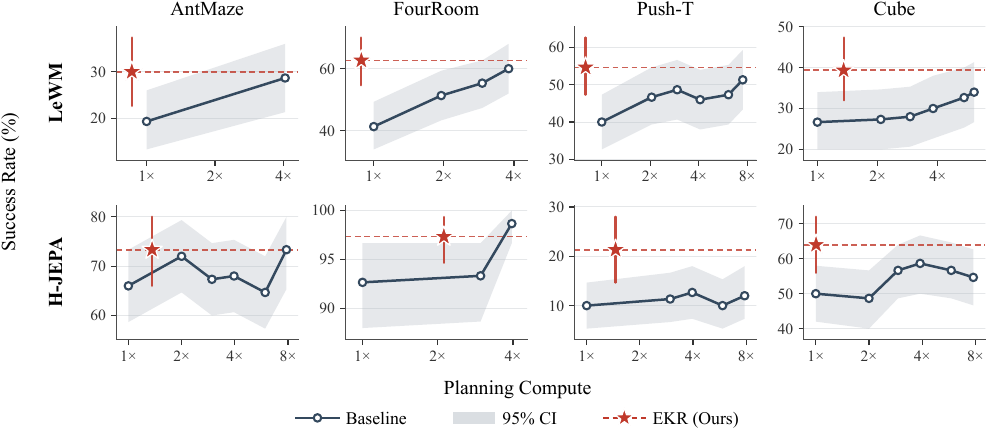}
\caption{\textbf{Success versus planning compute} (log scale, relative to
the $1\times$ baseline; \cref{app:metrics}). Line: baseline with
$1$--$8\times$ samples (shaded: 95\% bootstrap CI); star and dashed line:
\EKR{} at the default budget. All points use the same 150 (seed, task) pairs.}
\label{fig:budget}
\end{figure*}

\paragraph{Hierarchical world models.} For an $L$-level model with
level-$\ell$ encoder $E^{(\ell)}$ and predictor $F^{(\ell)}$, planning
proceeds top-down, and level $\ell=L,\dots,1$ solves
\begin{equation}
\begin{aligned}
  \mathbf{a}^{(\ell)\star}=\arg\min_{\mathbf{a}^{(\ell)}}\ &\Cz^{(\ell)}\big(\mathbf{a}^{(\ell)};g^{(\ell)}\big)\\
  &+\lambda_\ell\,\cRef^{(\ell)}\rho_K^{(\ell)}\big(\hz^{(\ell)}\big),
\end{aligned}
  \label{eq:hier}
\end{equation}
where $\hz^{(\ell)}$ is the level-$\ell$ predicted end latent and
$\lambda_\ell=0$ at levels without repulsion. The top level plans towards the
real goal; each lower level takes the first $S_{\ell+1}$ predicted latents of
the plan above as subgoals,
\begin{equation}
  g^{(L)}=E^{(L)}(o_g),\quad
  g^{(\ell)}=\big(\hz^{(\ell+1)\star}_{1},\dots,\hz^{(\ell+1)\star}_{S_{\ell+1}}\big),
  \label{eq:subgoal}
\end{equation}
where $\hz^{(\ell+1)\star}_{s}$ is the $s$-th latent rolled out by
$F^{(\ell+1)}$ under $\mathbf{a}^{(\ell+1)\star}$. $\Cz^{(L)}$ follows
\cref{eq:c0} in the level-$L$ space; for $\ell<L$ the goal term is measured
one level up,
\begin{equation}
  \Cz^{(\ell)}\big(\mathbf{a}^{(\ell)};g^{(\ell)}\big)=\sum_{s=1}^{S_{\ell+1}}\norm{\big[\Psi_{\ell+1}\big(\hz^{(\ell)}_{1:H_\ell}\big)\big]_s-g^{(\ell)}_s}_2^2,
  \label{eq:hier_c0}
\end{equation}
where $\Psi_{\ell+1}$ encodes level-$\ell$ predictions into level $\ell+1$ at
its temporal stride and $H_\ell$ is the level-$\ell$ horizon. The agent
executes the first action of $\mathbf{a}^{(1)\star}$. Each level with
$\lambda_\ell>0$ keeps its own memory of encodings $E^{(\ell)}(o_t)$ and
computes \cref{eq:dmin,eq:scale,eq:admission,eq:kernel,eq:rhok,eq:bandwidth,eq:cref}
in its own latent space, since geometry and cost magnitude differ across
levels. Upper-level repulsion changes the subgoals passed down and steers the
lower plans away from visited regions on a coarse time scale; lowest-level
repulsion suppresses revisits at the execution scale.

\paragraph{Use in a planner.} \EKR{} only replaces the cost function. In
\emph{gradient shaping} (our default for LeWM and H-JEPA), $\Ctot$ is
differentiated in every optimisation step and used for the final selection;
in \emph{selection-only} mode (Cube with H-JEPA), the optimiser runs on $\Cz$
and $\Ctot$ only re-ranks the final candidates; with a sampling planner such
as CEM (DINO-WM), $\Ctot$ ranks the samples in every iteration. The model,
samples, iterations, horizon and replanning schedule are unchanged, and
nothing is trained. The only extra cost is $\mathcal{O}(NM_td)$ distances per
cost evaluation, with no additional world-model rollouts.

\section{Experiments}
\label{sec:exp}

We ask whether \EKR{} improves success under the official protocol
(\cref{sec:main}), whether it is more than a larger search budget
(\cref{sec:compute}), which components matter (\cref{sec:ablation}), whether
memory causes the escape from traps (\cref{sec:behavior}), and whether it
transfers to other models and planners (\cref{sec:general}).

\subsection{Setup}
\label{sec:exp-setup}
\paragraph{Models and benchmarks.} We plan with two JEPA world models, using
their official weights and gradient-based MPC solvers: the flat \textbf{LeWM},
trained end-to-end from pixels~\citep{maes2026lewm}, and the three-level
\textbf{H-JEPA}, planned top-down through subgoals~\citep{zhang2026hjepa}.
The benchmarks are two navigation tasks whose walls and dead ends create
latent traps, Visual AntMaze~\citep{park2025ogbench} and
FourRoom~\citep{zhang2026hjepa}, and two short-horizon manipulation tasks,
Push-T~\citep{florence2021implicit,chi2023diffusion} and OGBench Cube, which
test whether \EKR{} also helps without obstacles.

\paragraph{Protocol.} Baseline and \EKR{} share the official tasks, success
criteria, budgets and planners, and differ only in the cost. Each
cell has 150 paired episodes (3 seeds $\times$ 50 tasks), compared with an
exact McNemar test~\citep{mcnemar1947note}. Hyperparameters are set per cell,
controlled by frozen reruns on held-out tasks and by one untuned
\emph{unified} configuration (\cref{app:hparams}). Compute counts executed
optimiser iterations relative to the baseline, including the repulsion
(\cref{app:metrics}).

\begin{table*}[t]
\centering
\caption{\textbf{Ablations and exploration controls} (success \%,
mean\,$\pm$\,std over 3 seeds $\times$ 50 tasks). Each row changes one
component of \EKR{}; the last block replaces the repulsion by other
exploration signals at the same budget. Count-based novelty: latent SimHash,
better of 8 or 4 bits per cell, which favours the control.}
\label{tab:ablation}
\setlength{\tabcolsep}{2.8pt}
\renewcommand{\arraystretch}{1.05}
\begin{tabular}{llcccccccc}
\toprule
 & & \multicolumn{2}{c}{FourRoom} & \multicolumn{2}{c}{Push-T} & \multicolumn{2}{c}{Cube} & \multicolumn{2}{c}{AntMaze} \\
\cmidrule(lr){3-4}\cmidrule(lr){5-6}\cmidrule(lr){7-8}\cmidrule(lr){9-10}
Group & Variant & LeWM & H-JEPA & LeWM & H-JEPA & LeWM & H-JEPA & LeWM & H-JEPA \\
\midrule
 & Baseline & \pmstd{41.3}{14.7} & \pmstd{92.7}{4.2} & \pmstd{40.0}{3.5} & \pmstd{10.0}{3.5} & \pmstd{26.7}{1.2} & \pmstd{50.0}{9.2} & \pmstd{19.3}{6.1} & \pmstd{66.0}{4.0} \\
 & \EKR{} (full) & \textbf{\pmstd{62.7}{5.8}} & \textbf{\pmstd{97.3}{2.3}} & \pmstd{54.7}{4.6} & \textbf{\pmstd{21.3}{5.0}} & \textbf{\pmstd{39.3}{7.6}} & \textbf{\pmstd{64.0}{12.0}} & \textbf{\pmstd{30.0}{9.2}} & \textbf{\pmstd{73.3}{9.5}} \\
\midrule
\multirow{2}{*}{\emph{\shortstack[l]{What to\\remember}}} & start state only & \pmstd{52.7}{11.4} & \pmstd{97.2}{0.0} & \pmstd{45.3}{7.0} & \pmstd{12.7}{4.2} & \pmstd{39.1}{8.1} & \pmstd{63.9}{12.1} & \pmstd{17.3}{4.2} & \pmstd{73.1}{6.4} \\
 & imagined states & \pmstd{39.3}{8.3} & \pmstd{96.0}{2.0} & \pmstd{42.0}{5.3} & \pmstd{7.3}{3.1} & \pmstd{25.3}{4.6} & \pmstd{52.0}{8.0} & \pmstd{14.7}{2.3} & \pmstd{69.3}{5.0} \\
\midrule
\multirow{2}{*}{\emph{\shortstack[l]{How to\\repel}}} & mean kernel (no top-$k$) & \pmstd{59.3}{8.1} & \pmstd{97.1}{2.0} & \textbf{\pmstd{56.7}{3.1}} & \pmstd{16.7}{2.3} & \textbf{\pmstd{39.3}{7.6}} & \pmstd{62.7}{11.4} & \pmstd{26.0}{10.4} & \pmstd{70.7}{6.1} \\
 & no $\cRef$ scaling & \pmstd{41.3}{9.5} & \pmstd{93.3}{1.2} & \pmstd{40.7}{8.3} & \pmstd{12.0}{2.0} & \pmstd{25.3}{6.1} & \pmstd{48.0}{6.0} & \pmstd{16.7}{1.2} & \pmstd{64.7}{11.5} \\
\midrule
\multirow{3}{*}{\emph{\shortstack[l]{Other\\signals}}} & count-based novelty & \pmstd{40.7}{13.6} & \pmstd{96.0}{4.0} & \pmstd{40.7}{4.2} & \pmstd{10.7}{8.1} & \pmstd{25.3}{1.2} & \pmstd{51.3}{6.1} & \pmstd{18.0}{6.0} & \pmstd{66.7}{3.1} \\
 & random cost (same scale) & \pmstd{16.0}{5.3} & \pmstd{74.7}{1.2} & \pmstd{12.7}{2.3} & \pmstd{10.0}{6.9} & \pmstd{39.0}{5.3} & \pmstd{38.0}{5.3} & \pmstd{21.3}{9.2} & \pmstd{47.3}{6.1} \\
 & action noise ($\epsilon{=}0.1$) & \pmstd{42.0}{12.0} & \pmstd{90.0}{3.5} & \pmstd{22.0}{6.0} & \pmstd{6.7}{4.2} & \pmstd{25.3}{4.2} & \pmstd{48.7}{11.0} & \pmstd{21.3}{8.1} & \pmstd{60.0}{10.4} \\
\bottomrule
\end{tabular}
\end{table*}

\subsection{Main results}
\label{sec:main}
\EKR{} improves success in all eight model--environment pairs, significantly
in six (\cref{tab:main}). On the flat LeWM, which has no subgoals to shorten
the horizon, all four gains are significant; on H-JEPA two are, so
hierarchical subgoals reduce latent traps but do not remove them. Tuning on
the evaluation tasks does not explain the gains: the frozen configurations
stay above the baseline in all eight held-out settings, and the untuned
unified configuration in seven of eight (\cref{tab:unified}).

\subsection{\EKR{} versus a larger search budget}
\label{sec:compute}
More samples also help, but at a much higher cost. In \cref{fig:budget}, a
star above or to the left of the baseline curve means that \EKR{} reaches the
same success with less compute. This holds in seven of eight settings, where
\EKR{} at no more than $1.5\times$ compute matches or exceeds $4\times$
samples; the exception is FourRoom/H-JEPA near the ceiling. The two levers differ in kind: more samples search the
same basin more densely, whereas the repulsion makes the visited basin costly
and moves the search out of it. On an exclusive A100 GPU, the repulsion adds ${\sim}25$\,ms per replanning
step ($<1\%$ of planning time) at unchanged GPU memory (\cref{app:metaworld}).

\subsection{Ablations and exploration controls}
\label{sec:ablation}
\Cref{tab:ablation} changes one component at a time (upper blocks) and
replaces the repulsion by other exploration signals (last block).

\paragraph{What to remember.} The memory must hold the states the agent has
actually visited. Remembering only the start state is significantly worse in
four settings and matches full memory only where the ceiling or a few
replans leave no room. Storing the
\emph{predicted} end points of earlier plans falls back to about the
baseline: imagined states inherit the model's errors, including the trap
itself, which is why \EKR{} stores observed latents (\cref{sec:memory}).

\paragraph{How to repel.} The aggregation matters little: a mean kernel over
the whole memory is indistinguishable from the local top-$k$ kernel in all
eight settings. The scale reference $\cRef$ (\cref{eq:cref}) is essential:
without it success returns to about the baseline everywhere, because the
magnitude of $\Cz$ varies across models and tasks and no fixed $\lambda$ fits
them all.

\paragraph{Other exploration signals.} Undirected exploration does not
replace memory. Action noise and count-based novelty stay near the baseline,
and a random cost of the same scale, the strongest control, is significantly
worse than \EKR{} in six of eight settings and often below the baseline.
What helps is not perturbing the plan but pushing it away from where the
agent has been.

\begin{figure*}[t]
\centering
\includegraphics[width=\textwidth]{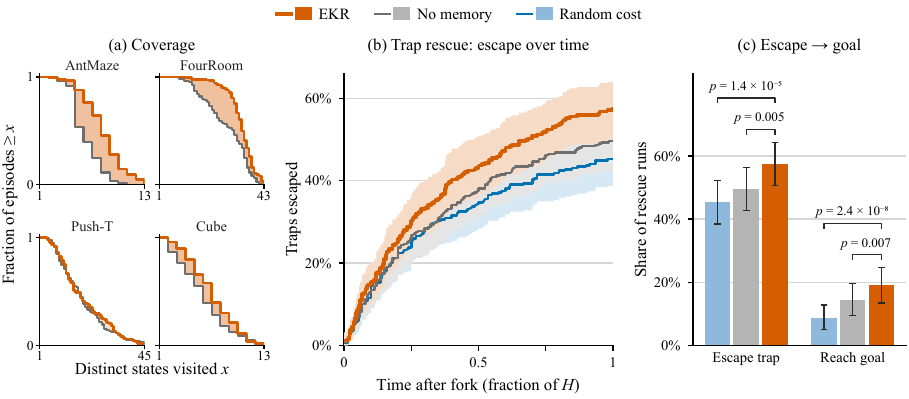}
\caption{\textbf{Memory changes behaviour in traps and causes the gain.}
(a)~Fraction of the 150 LeWM episodes of \cref{tab:main} that visit at least
$x$ distinct state cells; shading: \EKR{} ahead. On Push-T, \EKR{} episodes
end earlier because they reach the goal. (b)~Trap rescue on LeWM (160 traps):
baseline episodes are replayed into a trap and continued with identical seeds
under each cost; Kaplan--Meier fraction of traps left within $H$ steps
(shading: 95\% bootstrap band). (c)~Rescue runs that escape and that reach the
goal (95\% bootstrap CI; $p$: conditional logistic regression stratified by
trap).}
\label{fig:behavior}
\end{figure*}

\subsection{How memory changes behaviour}
\label{sec:behavior}
\paragraph{Behaviour in the main runs.} \EKR{} visits more distinct states
in all eight settings (curves further right in \cref{fig:behavior}a) and
spends less time trapped in seven, even in episodes that both methods fail,
so broader exploration is not a by-product of success (\cref{app:rescue}).

\begin{table}[b]
\centering
\caption{\textbf{Transfer to DINO-WM with CEM} (success \%, unified
configuration, mean\,$\pm$\,std over 3 planning seeds $\times$ 50 tasks;
budget in steps). $^{\dagger}$Pre-registered long-horizon split.
$\Delta$: absolute gain; exact McNemar test: $^{*}p{<}0.05$, $^{**}p{<}0.01$, $^{***}p{<}0.001$.}
\label{tab:general}
\setlength{\tabcolsep}{2pt}
\renewcommand{\arraystretch}{1.1}
\begin{tabular}{lccc}
\toprule
DINO-WM task (budget) & Baseline & + \EKR{} (ours) & $\Delta$ \\
\midrule
Wall, official (75) & \pmstd{90.0}{0.0} & \textbf{\pmstd{96.7}{3.1}}$^{**}$ & \up{6.7}{0} \\
Wall, long (150) & \pmstd{75.3}{4.2} & \textbf{\pmstd{94.7}{2.3}}$^{***}$ & \up{19.3}{0} \\
PointMaze, official (75) & \pmstd{100.0}{0.0} & \pmstd{100.0}{0.0} & 0.0 \\
PointMaze, long$^{\dagger}$ (150) & \pmstd{44.7}{4.2} & \textbf{\pmstd{100.0}{0.0}}$^{***}$ & \up{55.3}{0} \\
Push-T, official (75) & \pmstd{92.7}{4.2} & \textbf{\pmstd{96.7}{1.2}}$^{*}$ & \up{4.0}{0} \\
Push-T, long$^{\dagger}$ (150) & \pmstd{22.0}{3.5} & \textbf{\pmstd{33.3}{4.6}}$^{**}$ & \up{11.3}{0} \\
\bottomrule
\end{tabular}
\end{table}

\paragraph{A controlled rescue experiment.} To test whether the memory
\emph{causes} the escape, we take 160 trap moments from baseline LeWM runs,
replay each episode deterministically to a point deep inside the trap, and
continue it with identical state and planner seeds under three costs: the
baseline, \EKR{} with the agent's own memory, and the random cost
(\cref{app:rescue}). Switching the memory on makes the agent leave the trap
sooner (\cref{fig:behavior}b) and turns escapes into successes
(\cref{fig:behavior}c), raising the escape hazard by a factor of 1.45 and the
odds of reaching the goal by 2.80. The random cost lowers success, although
it reorders the candidate plans more often than \EKR{}. The gain thus comes
from repulsion anchored to \emph{real} visited states, not from perturbing
the plan ranking.

\subsection{Transfer to other models and planners}
\label{sec:general}
DINO-WM~\citep{zhou2025dinowm} plans on frozen DINOv2 features with CEM, so
it differs from our main models in both representation and optimiser. With
the official weights, tasks and planner and the unified configuration fixed
before any run (\cref{app:dinowm}), \EKR{} improves five of the six task sets
and leaves the sixth at the ceiling (\cref{tab:general}). The gain grows with
the horizon, most clearly on long PointMaze:
longer episodes offer more chances to fall into a trap. With a sampling
planner, \EKR{} acts mainly by shifting the CEM distribution over iterations
rather than by re-ranking the final samples. On the deeper hierarchies of
HWM~\citep{zhang2026hwm}, \EKR{} likewise improves all eight official cells,
three of them significantly (\cref{app:hwm-res}).

\section{Limitations}
\label{sec:limits}
\textbf{Escape is gradual.} Memory fills the current basin one entry per
replan, so deep dead ends require multiple steps to exit (median 53 steps on
AntMaze/LeWM); in short episodes with only about two replans, the plan is
unchanged.
\textbf{Goal-agnostic repulsion.} Because repulsion does not decay near the
goal, it can occasionally push the agent away from a visited near-goal state
(1 of 100 episodes on AntMaze/H-JEPA); annealing $\lambda$ with $\Cz$ is a
natural untested extension.
\textbf{Repulsion weight bounds.} While performance is stable across a
$16\times$--$32\times$ range in $\lambda$ and across $K$, $\tau_\sigma$ and $M$
(\cref{app:sensitivity}), excessive weights ($\lambda\ge 24$) overpower the
goal cost and degrade success. Further technical notes are in
\cref{app:limits}.

\section{Conclusion}
\label{sec:conclusion}
Goal-distance costs in latent world models induce traps where memoryless
receding-horizon planners stall far from the goal. \EKR{} turns the agent's
own observed trajectory into a scale-free, test-time repulsion signal without
training or extra search. Across the evaluated JEPA backbones, tasks and
planners (gradient-based MPC and CEM), it consistently reduces latent
revisits, matches or exceeds $4\times$ search budgets at ${\le}1.5\times$
compute and $<1\%$ wall-clock overhead, and transfers untuned to DINO-WM and
JEPA-WM. Episodic history is a simple, effective and broadly compatible
ingredient for world-model planning.

This paper presents work whose goal is to advance the field of machine
learning, specifically planning with learned world models. We do not foresee
societal consequences that must be specifically highlighted here.

\bibliography{refs}
\ificmlstyle
  \bibliographystyle{icml2026}  
\else
  \bibliographystyle{plainnat}
\fi

\clearpage
\appendix
\onecolumn

\section{Theoretical analysis}
\label{app:theory}
We analyse the final selection step (\cref{sec:fusion}). At replanning step
$t$ the agent is at the real latent $z_t$ with memory $\M_t$ (after the
admission decision for $z_t$), bandwidth $\sigma_t$, scale $c_t=\cRef>0$ and
final candidates $\mathcal{A}_t$:
\begin{equation}
  a_t^\star\in\arg\min_{a\in\mathcal{A}_t}\Ctot(a),\qquad
  \Ctot(a)=\Cz(a)+\lambda c_t\,\rho_K\big(\hz(a)\big),\qquad
  z_{t+1}=\Gamma(z_t,a_t^\star),
  \label{eq:th-select}
\end{equation}
where $\hz(a)$ is the predicted end latent and $\Gamma(z_t,a)$ the real latent
at the next replanning step, assumed to be determined by $z_t$ and $a$. For
$\mathcal{U}\subset\mathcal{Z}$ with $z_t\in\mathcal{U}$ and bounded
$\mathcal{V}\subset\mathcal{Z}$ we write
\begin{equation}
  \mathcal{A}_t^{\mathrm{stay}}=\{a\in\mathcal{A}_t:\Gamma(z_t,a)\in\mathcal{U}\},\qquad
  \mathcal{A}_t^{\mathrm{exit}}=\mathcal{A}_t\setminus\mathcal{A}_t^{\mathrm{stay}},
\end{equation}
\begin{align}
  \mathcal{P}(\mathcal{V},r)&=\max\big\{|X|:X\subseteq\mathcal{V},\ \norm{x-x'}_2>r\ \ \forall x\neq x'\in X\big\},\\
  \mathcal{N}(\mathcal{V},r)&=\min\big\{|Y|:Y\subset\mathcal{Z},\ \mathcal{V}\subseteq\textstyle\bigcup_{y\in Y}\bar B(y,r)\big\}.
\end{align}
A \emph{visit} to $\mathcal{U}$ is a maximal run $t_0,\dots,t_0+S$ with
$z_t\in\mathcal{U}$, $S\in\{0,1,\dots,\infty\}$; its $S$ steps with
$z_t,z_{t+1}\in\mathcal{U}$ are \emph{stay steps}, and $n_T(\mathcal{U})$
counts all stay steps before $T$.

\subsection{History-independent planners}
\label{app:theory-memoryless}
A planner is \emph{history-independent} if, for a fixed map $\Pi$,
\begin{equation}
  a_t^\star=\Pi(z_t,\omega_t),\qquad
  \omega_t\overset{\mathrm{i.i.d.}}{\sim}P_\omega\ \text{independent of } z_{0:t},\qquad
  q(z):=\Pr_\omega\big[\Gamma\big(z,\Pi(z,\omega)\big)\notin\mathcal{U}\big]
\end{equation}
($\Cz$ with any budget, action noise, terms coupling only current candidates;
a warm start adds the previous plan to the state). Then $(z_t)$ is Markov, and
a larger budget only changes $q$.

\begin{proposition}[History-independent planners]
\label{prop:memoryless}
If $\bar q=\sup_{z\in\mathcal{U}}q(z)<1$, every visit satisfies
\begin{equation}
  \Pr[S\ge n\mid S\ge n-1]\ge1-\bar q,\qquad
  \Pr[S\ge n]\ge(1-\bar q)^n,\qquad
  \mathbb{E}[S]=\sum_{n\ge1}\Pr[S\ge n]\ge\frac{1-\bar q}{\bar q},
  \label{eq:th-geom}
\end{equation}
and $S=\infty$ almost surely if $\bar q=0$ (\cref{eq:trap}).
\end{proposition}
\begin{proof}
By the Markov property, given the past and $z_{t_0+k-1}\in\mathcal{U}$, the
step stays in $\mathcal{U}$ with probability $1-q(z_{t_0+k-1})\ge1-\bar q$, so
\begin{equation}
  \Pr[S\ge n]=\prod_{k=1}^{n}\Pr[S\ge k\mid S\ge k-1]
  =\prod_{k=1}^{n}\mathbb{E}\big[1-q(z_{t_0+k-1})\,\big|\,S\ge k-1\big]\ge(1-\bar q)^n,
\end{equation}
and summing the geometric series gives $\mathbb{E}[S]$. The hazard thus stays
below $\bar q$ however long the agent has been in $\mathcal{U}$.
\end{proof}

\subsection{Bounded sojourn under \texorpdfstring{\EKR{}}{EKR}}
\label{app:theory-main}
At the steps with $z_t\in\mathcal{U}$ let $\varepsilon$ bound the gap between
predicted end latent and next real latent (model error plus the unexecuted
part of the plan), and assume, for constants $s_{\mathrm{lo}}>0$ and
$\gamma\ge1$ (typically $\gamma\approx1$ for the windowed median in
\cref{eq:scale}),
\begin{equation}
  \text{(A1)}\quad s_t\ge s_{\mathrm{lo}},\qquad s_{t+1}\le\gamma s_t,\qquad
  \varepsilon\ge\max_{a\in\mathcal{A}_t^{\mathrm{stay}}}\norm{\hz(a)-\Gamma(z_t,a)}_2;
\end{equation}
(A2)~some $a_e\in\mathcal{A}_t^{\mathrm{exit}}$ satisfies
\begin{equation}
  \Delta_t:=\frac{\Cz(a_e)-\min_{a\in\mathcal{A}_t^{\mathrm{stay}}}\Cz(a)}{c_t}
  <\lambda\Big(\rho_t^{\mathrm{lo}}-\rho_K\big(\hz(a_e)\big)\Big),\qquad
  \rho_t^{\mathrm{lo}}:=\frac1K\exp\!\Big(-\frac{(\gamma\alpha s_t+\varepsilon)^2}{2\sigma_t^2}\Big).
  \label{eq:th-margin}
\end{equation}
A trap of $\Cz$ has $\Delta_t>0$; (A2) bounds its depth by the repulsion gap
between remembered states and an exit far from the memory.

\begin{theorem}[Bounded sojourn]
\label{thm:sojourn}
Let $\mathcal{P}=\mathcal{P}(\mathcal{U},\alpha s_{\mathrm{lo}})$ and assume
(A1)--(A2). (i)~If no entry is evicted before $T$, then
$n_T(\mathcal{U})\le\mathcal{P}$. (ii)~If no entry is evicted during a visit,
or $M_{\max}\ge\mathcal{P}$, the visit has $S\le\mathcal{P}$, and
$S\le\mathcal{P}-1$ if its first latent $z_{t_0}$ is in memory throughout it.
Hence, under (ii), no visit makes more than $\mathcal{P}$ stay steps. Without
the gate (every real latent stored) this holds for any $\alpha>0$ in (A2).
\end{theorem}
\begin{proof}
Let $t$ be a stay step and suppose $d_{\min}(z_{t+1},\M_t)\le\alpha s_{t+1}$.
As $a_t^\star\in\mathcal{A}_t^{\mathrm{stay}}$, $d_{\min}$ is $1$-Lipschitz and
$\rho_K$ averages the $K_{\mathrm{eff}}\le K$ largest kernel values (so it is
at least $1/K$ times the largest), (A1)--(A2) give
\begin{align}
  d_{\min}\big(\hz(a_t^\star),\M_t\big)
  &\le d_{\min}(z_{t+1},\M_t)+\varepsilon\le\alpha s_{t+1}+\varepsilon\le\gamma\alpha s_t+\varepsilon,\\
  \rho_K\big(\hz(a_t^\star)\big)
  &\ge\frac1K\exp\!\Big(-\frac{d_{\min}(\hz(a_t^\star),\M_t)^2}{2\sigma_t^2}\Big)\ge\rho_t^{\mathrm{lo}},\\
  \Ctot(a_t^\star)
  &\ge\min_{a\in\mathcal{A}_t^{\mathrm{stay}}}\Cz(a)+\lambda c_t\,\rho_t^{\mathrm{lo}}
  \ \overset{\eqref{eq:th-margin}}{>}\ \Cz(a_e)+\lambda c_t\,\rho_K\big(\hz(a_e)\big)=\Ctot(a_e),
\end{align}
contradicting \cref{eq:th-select}. Hence $d_{\min}(z_{t+1},\M_t)>\alpha s_{t+1}$, and
$z_{t+1}$ is admitted (\cref{eq:admission}). Let $t_1<t_2<\cdots$ be stay steps
and $u_i=z_{t_i+1}\in\mathcal{U}$. If $u_i\in\M_{t_j}$ for all $i<j$, then
\begin{equation}
  \norm{u_j-u_i}_2\ \ge\ d_{\min}(u_j,\M_{t_j})\ >\ \alpha s_{t_j+1}\ \ge\ \alpha s_{\mathrm{lo}}
  \qquad(i<j),
\end{equation}
so the $u_i$ are $\alpha s_{\mathrm{lo}}$-separated in $\mathcal{U}$ and at most
$\mathcal{P}$ in number; so are $z_{t_0},u_1,u_2,\dots$ if also
$z_{t_0}\in\M_{t_j}$ for all $j$. Without eviction the premise holds. Within a
visit $t_{i+1}=t_i+1$, so $u_1,\dots,u_{j-1}$ are the last $j-1$ entries
written by step $t_j$; with $M_{\max}\ge\mathcal{P}$ they survive
first-in-first-out eviction for $j\le\mathcal{P}+1$, and $\mathcal{P}+1$ stay
steps would give $\mathcal{P}+1$ separated points.
\end{proof}

\paragraph{Generality.} All results but Prop.~\ref{prop:bandwidth} hold for any
kernel $k(\norm{\cdot}_2)$ with $k>0$ nonincreasing, $k(0)=1$, $k\to0$ (replace
$e^{-x^2/2\sigma_t^2}$ by $k(x)$), and any aggregation $\rho$ with
\begin{equation}
  \frac1K\max_{m\in\M_t}k\big(\norm{z-m}_2\big)\ \le\ \rho(z)\ \le\ \max_{m\in\M_t}k\big(\norm{z-m}_2\big);
\end{equation}
Theorem~\ref{thm:sojourn} holds for the mean kernel with $K\to|\M_t|$. The gate
is not needed for escape; as each entry is more than $\alpha s_{\min}$ from
every older entry still present, it bounds the memory independently of $M_{\max}$:
\begin{equation}
  |\M_t|\le\min\big\{M_{\max},\,\mathcal{P}(\mathcal{V},\alpha s_{\min})\big\}
  \qquad\text{for bounded }\mathcal{V}\supseteq\{z_\tau\}_{\tau\le t}.
\end{equation}

\begin{corollary}[Shape of the trap]
\label{cor:shape}
Let $r=\alpha s_{\mathrm{lo}}$. Then $\mathcal{P}(\mathcal{U},r)\le\mathcal{N}(\mathcal{U},r/2)$.
If $\mathcal{U}$ lies within distance $w<r/2$ of a rectifiable curve of length
$L>0$ (e.g.\ a tube around a cycle), resp.\ has diameter
$D_{\mathcal{U}}>r/2$ and doubling dimension $\mathfrak{d}$ ($\mathcal{U}\cap B(z,\varrho)$
is covered by $2^{\mathfrak{d}}$ balls of radius $\varrho/2$ for all $z$, $\varrho>0$), then
\begin{equation}
  \mathcal{P}(\mathcal{U},r)<\frac{L}{r-2w}+1,
  \qquad\text{resp.}\qquad
  \mathcal{P}(\mathcal{U},r)\le\mathcal{N}(\mathcal{U},r/2)<\Big(\frac{4D_{\mathcal{U}}}{r}\Big)^{\mathfrak{d}}.
  \label{eq:th-cycle}
\end{equation}
\end{corollary}
\begin{proof}
A closed ball of radius $r/2$ has diameter $r$, so it contains at most one
point of an $r$-separated set. Curve $\chi$: for separated $u_1,\dots,u_n$ pick
arc-length positions $\ell_1\le\dots\le\ell_n$ with
$\norm{u_i-\chi(\ell_i)}_2\le w$. Doubling: halve
$B(u_0,D_{\mathcal{U}})\supseteq\mathcal{U}$ ($u_0\in\mathcal{U}$) $j$ times,
with $j$ the smallest integer such that $D_{\mathcal{U}}2^{-j}\le r/2$
($j\ge1$ as $D_{\mathcal{U}}>r/2$). As a chord never exceeds the arc,
\begin{align}
  r<\norm{u_{i+1}-u_i}_2&\le\ell_{i+1}-\ell_i+2w
  &&\Longrightarrow\quad (n-1)(r-2w)<\textstyle\sum_{i<n}(\ell_{i+1}-\ell_i)\le L,\\
  \mathcal{N}(\mathcal{U},r/2)&\le2^{\mathfrak{d}j},\quad 2^{j-1}<\frac{2D_{\mathcal{U}}}{r}
  &&\Longrightarrow\quad \mathcal{N}(\mathcal{U},r/2)<\Big(\frac{4D_{\mathcal{U}}}{r}\Big)^{\mathfrak{d}},
\end{align}
which are the two bounds in \cref{eq:th-cycle}.
\end{proof}
The constants are the standard packing--covering ones ($2w$: two projection
errors; $4=2\cdot2$: radius $r/2$, rounding of $j$). A cycling trajectory
(\cref{sec:formulation}) is released after fewer than $L/(r-2w)+1$ stay
steps, independently of the latent dimension.

\begin{proposition}[Tightness]
\label{prop:tight}
For every $n\ge2$ there is an instance satisfying (A1)--(A2) with
$s_t\equiv s$, $\gamma=1$, $\varepsilon=0$, $M_{\max}\ge n$ and
$\mathcal{P}(\mathcal{U},\alpha s)=n$ in which the visit from $z_0$ has exactly
$n-1$ stay steps, attaining $S\le\mathcal{P}-1$ in Theorem~\ref{thm:sojourn}.
\end{proposition}
\begin{proof}
With $\sigma_t\equiv\sigma$ and $\rho^{\mathrm{lo}}=\frac1Ke^{-\alpha^2s^2/2\sigma^2}$, fix in turn
\begin{equation}
\begin{cases}
  \delta: & e^{-\delta^2/2\sigma^2}<\rho^{\mathrm{lo}},\\[2pt]
  D: & D>\alpha s,\quad \lambda e^{-D^2/2\sigma^2}<\min\big\{\lambda(\rho^{\mathrm{lo}}-e^{-\delta^2/2\sigma^2}),\,n+1\big\},\\[2pt]
  \Delta: & \lambda e^{-D^2/2\sigma^2}<\Delta<\min\big\{\lambda(\rho^{\mathrm{lo}}-e^{-\delta^2/2\sigma^2}),\,n+1\big\}.
\end{cases}
\end{equation}
In $\mathbb{R}^3$ put $\mathcal{U}=\{u_1,\dots,u_n\}$
on a circle about $z_g$ of radius $R\ge\max\{\delta,D/(2\sin\frac{\pi}{n})\}$ with spacing $\ge D$,
$z_0=u_1$, and $z_e$ on its axis with $\norm{z_e-z_g}_2^2=R^2+c_t\Delta$, so
$\mathrm{dist}(z_e,\mathcal{U})>\delta$. Error-free candidates $a_j\mapsto u_j$
and $a_e\mapsto z_e$, scored by squared end-point goal distances ($B=1$), have
\begin{equation}
  \Cz(a_j)=b:=R^2,\qquad \Cz(a_e)=b+c_t\Delta,\qquad
  c_t=\frac{nb+(b+c_t\Delta)}{n+1}=\frac{(n+1)b}{n+1-\Delta},
\end{equation}
so \cref{eq:trap} and (A2) hold. Every reached $u_j$ is admitted, and for an
unvisited $u_j$ and a visited $u_k$,
\begin{equation}
  \Ctot(a_j)\le b+\lambda c_te^{-D^2/2\sigma^2}<b+c_t\Delta\le\Ctot(a_e)
  <b+\lambda c_t\rho^{\mathrm{lo}}\le b+\tfrac{\lambda c_t}{K}\le\Ctot(a_k).
\end{equation}
\EKR{} therefore visits all $n$ points, keeping $z_0$ in memory, and then leaves.
\end{proof}
The slow escape in \cref{sec:limits} is thus intrinsic: \EKR{} prefers the
unvisited states of a trap to a deeper exit.

\begin{proposition}[Bandwidth window and depth capacity]
\label{prop:bandwidth}
(A2) forces $\Delta_t<\lambda/K$. For $r=\gamma\alpha s_t+\varepsilon$ and an
exit at distance at least $\delta$ from the memory, $\theta=r/\delta\in(0,1)$, the
margin in \cref{eq:th-margin} is at least $\lambda g(\sigma_t)$; $g>0$ for all
$\sigma$ if $K=1$, $\lim_{\sigma\to0}g=0$, $\lim_{\sigma\to\infty}g=\frac1K-1$, and
\begin{align}
  g(\sigma)&=\frac1Ke^{-r^2/2\sigma^2}-e^{-\delta^2/2\sigma^2},&
  g(\sigma)>0\ &\Longleftrightarrow\ \sigma^2<\frac{\delta^2-r^2}{2\ln K}\quad(K>1),\\
  \sigma_\star^2&=\frac{\delta^2-r^2}{2\ln(K/\theta^2)}=\arg\max g,&
  \frac{g(\sigma_\star)}{1/K}&=(1-\theta^2)\Big(\frac{\theta^2}{K}\Big)^{\theta^2/(1-\theta^2)}
  \xrightarrow{\ \theta\to0\ }1.
  \label{eq:th-sigma}
\end{align}
\end{proposition}
\begin{proof}
$\rho_t^{\mathrm{lo}}\le1/K$ and $\rho_K\ge0$ give the first claim, and $g>0$
is equivalent to $(\delta^2-r^2)/(2\sigma^2)>\ln K$. With $v=1/(2\sigma^2)$,
\begin{align}
  \partial_vg&=e^{-r^2v}\big(\delta^2e^{-(\delta^2-r^2)v}-r^2/K\big)\gtrless0
  \ \Longleftrightarrow\ v\lessgtr v_\star,\qquad e^{(\delta^2-r^2)v_\star}=K/\theta^2,\\
  g(\sigma_\star)&=\frac1Ke^{-r^2v_\star}-e^{-\delta^2v_\star}
  =\frac{1-\theta^2}{K}\,e^{-r^2v_\star},\qquad
  e^{-r^2v_\star}=\Big(\frac{\theta^2}{K}\Big)^{\theta^2/(1-\theta^2)},
\end{align}
which gives $\sigma_\star$ and $g(\sigma_\star)$ in \cref{eq:th-sigma}.
\end{proof}
The optimum attains almost the full capacity $\lambda/K$ ($98\%$ for
$\theta=0.05$, $K=3$). As $\sigma_\star$ depends on the unknown $\delta$,
\EKR{} ties $\sigma$ to the spacing ($\sigma_t=s_t$) or to the candidates
(\cref{eq:bandwidth}); the worst-case factor $1/K$ does not predict the choice of $K$.

\subsection{The random-cost control and the window for \texorpdfstring{$\lambda$}{lambda}}
\label{app:theory-random}
The control in \cref{tab:ablation} replaces $\rho_K(\hz(a))$ by
$\mu_t+\sigma^\rho_t\zeta_t(a)$, with $\mu_t,\sigma^\rho_t$ the mean and
standard deviation of $\rho_K$ over the candidates and $\zeta_t$ the
candidate-standardised ($0$ if degenerate) field
\begin{equation}
  f_t(z)=\frac12+\frac1{2J}\sum_{j=1}^{J}\sin\big(\xi_j^\top z+\vartheta_j\big),\qquad
  \xi_j\sim\mathrm{Normal}(0,\varsigma_t^{-2}I),\quad \vartheta_j\sim\mathrm{Unif}[0,2\pi),
  \label{eq:th-field}
\end{equation}
at $z=\hz(a)$, with $J=16$, $\varsigma_t$ the spread of the end latents and a
fresh draw per replanning step: it matches the repulsion in mean, spread,
smoothness and weight, not in direction. Candidates are taken as given and
independent of the draw (exact for selection, idealised for gradient shaping).

\begin{proposition}[Sign symmetry and locality]
\label{prop:random}
(a)~$(\zeta_t(a))_{a\in\mathcal{A}_t}\overset{d}{=}(-\zeta_t(a))_{a\in\mathcal{A}_t}$,
and this law does not depend on $\M_t$. Hence, if $\Cz(a')-\Cz(a)=\eta c_t$
with $\eta\ge0$, $\sigma^\rho_t>0$, and $a'\succ a$ denotes a strictly lower
randomised cost of $a'$,
\begin{equation}
  \mathbb{E}\big[\zeta_t(a)-\zeta_t(a')\big]=0,\qquad
  \Pr[a'\succ a]=\tfrac12\Pr\Big[|\zeta_t(a)-\zeta_t(a')|>\frac{\eta}{\lambda\sigma^\rho_t}\Big]\le\tfrac12 .
  \label{eq:th-coin}
\end{equation}
(b)~If $\hz(a),\hz(a')$ are at least $d_0$ from $\M_t$, then
\begin{equation}
  \Big|\big[\Ctot(a)-\Ctot(a')\big]-\big[\Cz(a)-\Cz(a')\big]\Big|\le\lambda c_t\,e^{-d_0^2/2\sigma_t^2}.
\end{equation}
\end{proposition}
\begin{proof}
(a)~The shift $\vartheta_j\mapsto\vartheta_j+\pi\pmod{2\pi}$ preserves the law of
$(\xi_j,\vartheta_j)$, and $\varsigma_t$ depends only on the candidates. With
$X:=\zeta_t(a)-\zeta_t(a')$,
\begin{equation}
  f_t-\tfrac12\ \mapsto\ -\big(f_t-\tfrac12\big),\qquad
  \zeta_t\ \mapsto\ -\zeta_t,\qquad
  X\overset{d}{=}-X,\qquad
  a'\succ a\ \Longleftrightarrow\ X>\frac{\eta}{\lambda\sigma^\rho_t},
\end{equation}
so $X$ is symmetric and bounded, $\mathbb{E}X=0$, and
$\Pr[a'\succ a]=\frac12\Pr[|X|>\eta/(\lambda\sigma^\rho_t)]$. (b)~Both
$\rho_K$ values lie in $[0,e^{-d_0^2/2\sigma_t^2}]$.
\end{proof}
In a trap ($a$ the best stay, $a'$ an exit) the random cost lets the exit beat
the best stay with probability at most $\frac12$ per step, the repulsion with
certainty once the trap is covered (Theorem~\ref{thm:sojourn}). Away from traps
($a,a'$ the two best under $\Cz$, $\eta\ll\lambda\sigma^\rho_t$) it flips the
choice with probability near $\frac12$; by (b) the repulsion does not once
$\eta>\lambda e^{-d_0^2/2\sigma_t^2}$.

\begin{proposition}[Two-sided window]
\label{prop:window}
For fixed step data at $t$ (candidates, memory, $\sigma_t$), let $a_g$ (e.g.\ a
goal-reaching candidate) satisfy
\begin{equation}
  \Delta_t^g:=\frac{\min_{a\neq a_g}\Cz(a)-\Cz(a_g)}{c_t}>0,\qquad
  d_g:=d_{\min}\big(\hz(a_g),\M_t\big);\qquad\text{then}\quad
  \lambda<\Delta_t^g\,e^{d_g^2/2\sigma_t^2}\ \Longrightarrow\ a_t^\star=a_g.
\end{equation}
Hence, if $\rho_t^{\mathrm{lo}}>\rho_K(\hz(a_e))$ at
the trap steps $\mathcal{I}_{\mathrm{trap}}$ of a trajectory ($a_e$ as in (A2)),
(A2) there and $a_t^\star=a_g$ at its goal steps $\mathcal{I}_{\mathrm{goal}}$
hold whenever $\lambda\in\Lambda$, with $\Lambda$ evaluated on that trajectory:
\begin{equation}
  \Lambda:=\Big(\sup_{t\in\mathcal{I}_{\mathrm{trap}}}\frac{\Delta_t}{\rho_t^{\mathrm{lo}}-\rho_K(\hz(a_e))},\
  \inf_{t\in\mathcal{I}_{\mathrm{goal}}}\Delta_t^g\,e^{d_g^2/2\sigma_t^2}\Big),
  \qquad
  \Lambda\big[c\,\Cz\big]=\Lambda\big[\Cz\big]\quad\forall c>0.
  \label{eq:th-window}
\end{equation}
\end{proposition}
\begin{proof}
For every $a\neq a_g$, since $\rho_K(\hz(a_g))\le e^{-d_g^2/2\sigma_t^2}$ and $\rho_K\ge0$,
\begin{equation}
  \Ctot(a_g)\le\Cz(a_g)+\lambda c_te^{-d_g^2/2\sigma_t^2}<\Cz(a_g)+c_t\Delta_t^g\le\Cz(a)\le\Ctot(a).
\end{equation}
The lower end is \cref{eq:th-margin}. For fixed step data, $\Delta_t$,
$\Delta_t^g$ and $\rho_K$ are invariant under $\Cz\mapsto c\,\Cz$ by
\cref{eq:invariance}.
\end{proof}
A goal-reaching candidate is displaced only once the goal region is in memory
($d_g\lesssim\sigma_t$) with margin below $\lambda$. A constant $\bar c$ instead
of $c_t$ scales all endpoints by $c_t/\bar c$, which varies across backbones
and episodes; with $\cRef$, $\Lambda$ is dimensionless.

\subsection{Discussion}
\label{app:theory-disc}
With $\sigma_t=s_t$, $\gamma=1$, $\varepsilon=0$ and an exit at distance
$\delta=\nu s_t$ from the memory, \cref{eq:th-margin} is implied by
\begin{equation}
  \Delta_t<\lambda\Big(\frac{e^{-\alpha^2/2}}{K}-e^{-\nu^2/2}\Big),
  \qquad\text{which is nonvacuous iff}\quad \nu>\sqrt{2\ln K+\alpha^2}.
\end{equation}
For FourRoom/LeWM ($\alpha=0.1$, $\lambda=2$, $K=3$; \cref{tab:hparams}),
$\nu>1.49$, and the bound covers exits costlier than the best stay by up to
$0.39\,c_t$ at $\nu=2$ and $0.64\,c_t$ at $\nu=3$, near the capacity
$\lambda/K=0.67$; a gap $\varepsilon=s_t/2$ still gives
$\rho_t^{\mathrm{lo}}=0.84/K$. The analysis assumes, but does not show, that
the planner proposes such an exit; it applies to CEM when the best-ranked
sample is executed, and to a hierarchical level whose selected plan determines
the next real latent up to $\varepsilon$.

\begin{table}[H]
\centering
\caption{Theoretical results and the observations they account for.}
\label{tab:th-exp}
\footnotesize
\begin{tabular}{@{}p{0.43\textwidth}p{0.15\textwidth}p{0.36\textwidth}@{}}
\toprule
Observation & Result & Explanation \\
\midrule
Budget scaling is costly (\cref{fig:budget}); noise $\approx$ baseline & Prop.~\ref{prop:memoryless} & Hazard $\le\bar q$ at all times; budget only changes $\bar q$. \\
Dead ends left slowly; few replans give no effect (\cref{sec:limits}) & Thm.~\ref{thm:sojourn}, Prop.~\ref{prop:tight} & One entry per stay step; bound attained. \\
Imagined states lose the gain & Thm.~\ref{thm:sojourn} & The count needs reached latents in $\mathcal{U}$. \\
Start-only loses gain, except Cube (${\approx}4$ replans) & (A2), Thm.~\ref{thm:sojourn} & Distant stays not repelled; memory needs steps. \\
No gate: no significant change & Thm.~\ref{thm:sojourn} & Escape does not need the gate. \\
No $\cRef$ removes the gain & Prop.~\ref{prop:window} & Endpoints of $\Lambda$ move with $c_t/\bar c$. \\
Random cost is below the baseline in five settings & Prop.~\ref{prop:random} & No preference for exits; perturbs far choices. \\
Rare reached-then-pushed-away failures (\cref{sec:limits}) & Prop.~\ref{prop:window} & Needs goal region in memory, margin $<\lambda$. \\
Mean kernel performs like top-$k$ & Thm.~\ref{thm:sojourn} (weak) & Holds with $K\to|\M_t|$; factor $1/K$ is loose. \\
DINO-WM: selection is usually the $\Cz$-best (\cref{sec:general}) & not covered & \EKR{} acts through the CEM distribution. \\
\bottomrule
\end{tabular}
\end{table}

\section{Hyperparameters and configuration disclosure}
\label{app:hparams}

\paragraph{Shared harness.} Official \texttt{eval.py} configuration for all
eight cells (50 evaluation tasks; episode budgets of 350, 210, 90 and 30
steps on AntMaze, FourRoom, Push-T and Cube; \texttt{chunk\_size}=1). Deterministic execution (fixed environment and
planner seeds, deterministic cuBLAS, TF32 off, two threads per process).
Seeds 42/43/44 correspond to the official pretrained weights of the same
seed. The planner (gradient-based MPC, AdamW, learning rate 0.05, up to 90
iterations with early stopping after at least 30 iterations and patience 5;
horizon and receding schedule as released) is unchanged by \EKR{}. Sample
counts per replanning step are given in \cref{tab:samples}. Within each
cell, baseline and \EKR{} share the same software environment.

\begin{table}[h]
\centering
\caption{Official number of planner samples per replanning step (L1--L3:
H-JEPA hierarchy levels).}
\label{tab:samples}
\begin{tabular}{lcc}
\toprule
Environment & LeWM & H-JEPA (L1/L2/L3) \\
\midrule
AntMaze  & 100 & 100 / 100 / 400 \\
FourRoom & 200 & 50 / 200 / 200 \\
Push-T   & 800 & 25 / 50 / 200 \\
Cube     & 800 & 25 / 800 / 800 \\
\bottomrule
\end{tabular}
\end{table}

\begin{table}[h]
\centering
\caption{\textbf{\EKR{} configuration per cell.} $M_{\max}$: memory capacity;
$\alpha$: novelty coefficient of \cref{eq:admission}. ``obs'': observation calibration $\sigma=s_t$
(twice the median real-trajectory spacing, $\beta{=}2$);
``endpoint'': endpoint calibration \eqref{eq:bandwidth}. All cells use $W{=}64$ and $\beta{=}2$ (Cube: $\beta{=}1$) in \cref{eq:scale}.}
\label{tab:hparams}
\setlength{\tabcolsep}{4pt}
\begin{tabular}{llcccccll}
\toprule
Model & Env & $\lambda$ & $K$ & levels & $\sigma$ mode & $M_{\max}$ & $\alpha$ & where applied \\
\midrule
LeWM   & AntMaze  & 1.2 & 3 & -- & obs ($\times2$) & 256 & 0.1 & gradient + selection \\
LeWM   & FourRoom & 2.0 & 3 & -- & obs ($\times2$) & 256 & 0.1 & gradient + selection \\
LeWM   & Push-T   & 1.0 & 5 & -- & obs ($\times2$) & 256 & 0.1 & gradient + selection \\
LeWM   & Cube     & 0.5 & 1 & -- & endpoint ($\tau_\sigma{=}1.5$) & 128 & 0.1 & gradient + selection \\
H-JEPA & AntMaze  & 1.0 & 3 & L3 & endpoint ($\tau_\sigma{=}1$) & 128 & 0.1 & gradient + selection \\
H-JEPA & FourRoom & 1.5 & 3 & L1--L3 & endpoint ($\tau_\sigma{=}1$) & 128 & 0.1 & gradient + selection \\
H-JEPA & Push-T   & 1.5 & 5 & L1--L3 & endpoint ($\tau_\sigma{=}1$) & 128 & 0.1 & gradient + selection \\
H-JEPA & Cube     & 10 / 25 & 1 & L1, L2 & endpoint ($\tau_\sigma{=}1.5$) & 128 & 0.1 & selection only \\
\bottomrule
\end{tabular}
\end{table}

\paragraph{Notes.}
(i) For LeWM on AntMaze/FourRoom/Push-T, $\cRef$ is computed once per
replanning step from the first gradient iteration and held fixed within the
step; elsewhere it is recomputed at every cost evaluation. Both are the
stop-gradient mean base cost of \cref{eq:total}.
(ii) For Cube/H-JEPA the repulsion is applied only in the final selection of
the best candidate, with per-level strengths ($\lambda_{L1}{=}10$,
$\lambda_{L2}{=}25$). A z-scored variant of the scale factor reached 95/150
successes versus 96/150 for the reported $\cRef$ variant (paired 21/20,
$p{=}1.0$).
(iii) The configurations above are frozen and re-evaluated unchanged on a
held-out task set (\cref{tab:main}; 50 new tasks per seed drawn from the
official distribution with task seed 10042), where the gain stays positive in
all eight cells. The unified configuration ($K{=}3$,
$\lambda{=}1.5$, all levels, endpoint $\tau_\sigma{=}1$) coincides with the
per-cell one on FourRoom/H-JEPA (97.3\%) and is used without modification
for every new backbone and environment in \cref{sec:general}.

\paragraph{HWM configurations.}\label{app:hwm} HWM-L3 uses a per-cell configuration on AntMaze, Push-T and Cube and an
untuned one on FourRoom. For HWM-L2, AntMaze reuses the HWM-L3
configuration, FourRoom is untuned, and Push-T and Cube have their own
configurations.

\paragraph{DINO-WM protocol.}\label{app:dinowm} We use the official weights,
tasks and CEM planner (300 samples, horizon 5, 30 optimisation steps).
Because the official MPC replans without limit, we set a finite execution
budget of three times the median baseline solve time, fixed from baseline
runs before any \EKR{} run. \EKR{} uses the unified configuration (endpoint
bandwidth with $\tau_\sigma{=}1$, memory capacity 128). The long Push-T split
uses CEM horizon 10 and 10 actions per replan for both arms. The selected
sample is usually also the best one under $\Cz$ (median rank 0), so \EKR{}
acts mainly through the CEM sampling distribution. On the long Wall tasks
this costs 3--15 extra steps when both arms succeed; on long PointMaze
\EKR{} is faster (about 36 vs.\ 51 steps). Paired \EKR{}-only/baseline-only
successes in \cref{tab:general} are 11:1 (Wall, official), 29:0 (Wall, long),
83:0 (PointMaze, long; $p{=}2\times10^{-25}$), 6:0 (Push-T, official;
$p{=}0.031$) and 23:6 (Push-T, long; $p{=}0.0023$).

\begin{table}[h]
\centering
\caption{\textbf{Unified configuration on the eight main cells} (success \%,
3 seeds $\times$ 50 tasks; $K{=}3$, $\lambda{=}1.5$, all levels, endpoint
$\tau_\sigma{=}1$; no per-cell tuning). Exact McNemar $p$ on 150 paired
episodes.}
\label{tab:unified}
\setlength{\tabcolsep}{6pt}
\small
\begin{tabular}{llcccc}
\toprule
Env. & Backbone & Baseline & Unified & $\Delta$ & $p$ \\
\midrule
AntMaze  & LeWM    & 19.3 & 26.0 & $+6.7$  & 0.11 \\
AntMaze  & H-JEPA  & 66.0 & 75.3 & $+9.3$  & 0.076 \\
FourRoom & LeWM    & 41.3 & 66.7 & $+25.3$ & $3.2\times10^{-8}$ \\
FourRoom & H-JEPA  & 92.7 & 97.3 & $+4.7$  & 0.12 \\
Push-T   & LeWM    & 40.0 & 47.3 & $+7.3$  & 0.16 \\
Push-T   & H-JEPA  & 10.0 & 18.7 & $+8.7$  & 0.047 \\
Cube     & LeWM    & 26.7 & 34.7 & $+8.0$  & 0.11 \\
Cube     & H-JEPA  & 50.0 & 48.7 & $-1.3$  & 0.89 \\
\bottomrule
\end{tabular}
\end{table}

\Cref{tab:unified} shows the unified configuration on all eight main cells. It
is above the baseline in seven and significant in two (FourRoom/LeWM and
Push-T/H-JEPA); on Cube/H-JEPA, where \EKR{} is selection-only, it gives no
gain. The per-cell configurations of \cref{tab:main} give larger gains in most
cells, most clearly on Cube/H-JEPA, so part of the main-table gain comes from
tuning.

\begin{table}[h]
\centering
\caption{Per-seed successes (out of 50) and paired win/loss counts
(\EKR{} succeeds and baseline fails / vice versa) over 150 episodes.}
\label{tab:wins}
\setlength{\tabcolsep}{5pt}
\begin{tabular}{llccc}
\toprule
Model & Env & Baseline (seeds 42/43/44) & \EKR{} (seeds 42/43/44) & wins / losses ($p$) \\
\midrule
LeWM   & AntMaze  & 7 / 9 / 13   & 20 / 14 / 11 & 33 / 17 (0.033) \\
LeWM   & FourRoom & 15 / 29 / 18 & 28 / 33 / 33 & 41 / 9 ($5.6\times10^{-6}$) \\
LeWM   & Push-T   & 21 / 21 / 18 & 26 / 30 / 26 & 38 / 16 (0.0038) \\
LeWM   & Cube     & 13 / 14 / 13 & 18 / 17 / 24 & 34 / 15 (0.0094) \\
H-JEPA & AntMaze  & 33 / 35 / 31 & 42 / 35 / 33 & 33 / 22 (0.18) \\
H-JEPA & FourRoom & 44 / 48 / 47 & 48 / 50 / 48 & 11 / 4 (0.12) \\
H-JEPA & Push-T   & 7 / 4 / 4    & 13 / 11 / 8  & 25 / 8 (0.0046) \\
H-JEPA & Cube     & 29 / 26 / 20 & 26 / 32 / 38 & 46 / 25 (0.017) \\
\bottomrule
\end{tabular}
\end{table}

\section{Protocol details}
\label{app:metrics}
\paragraph{Reproducibility.} Every main-table cell (baseline and \EKR{},
3 seeds $\times$ 50 tasks) was run twice; all 16 cells agree
episode-by-episode (identical actions, step counts and outcomes; at most
floating-point noise $\le 4\times10^{-6}$ in a few recorded states).
Per-episode comparison videos (reference / executed / goal) were exported for
all 2400 main-table episodes.

\paragraph{Compute.} Relative compute in \cref{fig:budget} is the ratio of
executed gradient iterations (both arms $\ge$20 episodes), otherwise the
nominal multiplier ($2\times$/$4\times$ samples); for
\EKR{} it is multiplied by $(1+f)$, where $f$ is the fraction of planning
time spent on the repulsion. Budgets whose runs do not cover all three seeds are omitted from
\cref{fig:budget}. Clean wall-clock overhead measured on an exclusive A100 GPU
is ${\sim}25$\,ms per replan ($<1\%$ of planning time) at unchanged peak GPU
memory (\cref{app:metaworld}).
Capping the memory at 8 entries roughly halves the repulsion time on
AntMaze/LeWM in a separate run but lowers success from 20 to 15 of 50 (baseline 7;
seed 42), so we keep the default capacity.

\section{Behaviour analysis and trap-rescue experiment}
\label{app:rescue}

\paragraph{Trap criterion.} Let $d(t)$ be the task distance normalised by
$d(0)$ and $b(t)=\min_{s\le t}d(s)$. Step $t$ of an unfinished episode is
\emph{trapped} if $b$ improved by less than $0.05$ over the last $W$ steps,
with $W=30$ (FourRoom), 20 (Push-T), 10 (Cube) and 50 (AntMaze).

\paragraph{Behaviour in the main runs.} Coverage is the number of distinct
cells of the true state visited in an episode (AntMaze 2.0 units, FourRoom
8\,px, Push-T block 16\,px $\times\,\pi/8$, Cube 2\,cm); we report the ratio of
the \EKR{} and baseline means over the 150 paired episodes per setting. Trapped
time is the fraction of trapped steps within the first $\min(T_{\text{base}},
T_{\EKR{}})$ steps of each pair, so that shorter successful episodes do not
favour either method. Tests are paired Wilcoxon tests with Holm correction over
the eight settings; CIs are paired bootstrap (2000 resamples).

\paragraph{Rescue protocol.} For each LeWM benchmark we ran the baseline with
the main-table configuration on seeds 42--44 (AntMaze: tasks 0--23) and took
the first trap of every episode that had not yet escaped and had at least $2W$
steps of budget left after the fork; 40 such traps per benchmark were drawn at
random (AntMaze/H-JEPA: 40 more, same procedure). Each episode is replayed with
its recorded actions to the first replanning step after detection, i.e.\ $W$
steps into the trap; the replay reproduces the physical state exactly (maximum
deviation $0$), while the planner and the memory run as in the original
episode. From there the episode continues in branches with identical
state and reseeded planner: \emph{base} (no repulsion), \emph{own} (\EKR{} with
the configuration of \cref{tab:main} and the memory collected up to the fork),
and \emph{rand} (the random cost of
\cref{tab:ablation}, matched in mean and spread). Each branch is run with two
planner seeds whose outcomes are averaged per trap. The primary outcome is
escape within $H=\min(3W,\text{remaining budget})$ steps: the normalised
distance falls more than 0.05 below its best value at trap onset, or the task
is solved. Secondary outcomes are the escape time and success from the trap
(by the end of the official budget). Branches, outcomes, sampling, sample sizes
and tests were fixed before the main runs, and all benchmarks are reported. These
conditional outcomes are not success rates of \cref{tab:main}; their base rates
depend on the sampled traps.

\begin{figure}[htbp]
\centering
\includegraphics[width=0.38\textwidth]{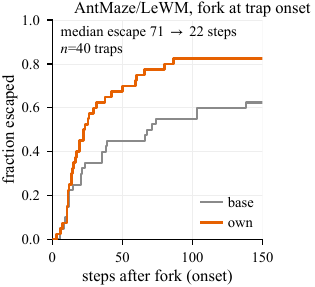}
\caption{\textbf{Forking at trap onset on AntMaze/LeWM.} Escape fraction
within $H$ steps when forking at trap detection rather than $W$ steps into the
trap ($n{=}40$ traps; escape $0.62\to 0.82$, $p{=}0.02$; median escape time
$71\to 22$ steps).}
\label{fig:rescue-onset}
\end{figure}

\paragraph{Results.} \Cref{tab:rescue-cells} gives per-benchmark results and
\cref{tab:rescue-pooled} the pooled models. The effect is largest on the navigation tasks,
where an episode accumulates many memory entries before the trap (FourRoom: escape
time hazard ratio 1.99, success from the trap 0.20 to 0.38; AntMaze: hazard ratio
1.82). On Push-T the memory does not raise the escape rate over the baseline
but is far better than the random cost, which lowers it from 0.70 to 0.55. On Cube
the memory holds only about two entries at the trap (four replans per episode),
the short-horizon case in which \EKR{} and any perturbation behave alike
(\cref{sec:ablation}). On AntMaze/H-JEPA the baseline already leaves 86\% of the
traps within $H$, leaving no room for improvement; including it in the pooled
analysis leaves the conclusions unchanged (\cref{tab:rescue-pooled}).
Forking at trap detection instead of $W$ steps later also accelerates escape
on AntMaze/LeWM ($0.62$ to $0.82$, $p{=}0.02$, median escape time $71$ to $22$
steps; \cref{fig:rescue-onset}).

\begin{table*}[t]
\centering
\caption{\textbf{Trap rescue per benchmark} (40 traps each, two planner seeds
per trap and branch). Escape within $H$ and success from the trap are trap-level
means; own$-$base: paired difference with bootstrap CI and Wilcoxon $p$, Holm
corrected over the five rows; HR: escape-time hazard ratio from a Cox model
stratified by trap. $^\dagger$Ceiling: the baseline leaves 86\% of the traps.}
\label{tab:rescue-cells}
\setlength{\tabcolsep}{3.5pt}
\small
\begin{tabular}{l c ccc c c c ccc}
\toprule
 & & \multicolumn{3}{c}{escape@$H$} & & & & \multicolumn{3}{c}{success from trap}\\
\cmidrule(lr){3-5}\cmidrule(lr){9-11}
Cell & $n$ & \textit{base} & \textit{own} & \textit{rand} & \textit{own}$-$\textit{base} [95\% CI] & $p_\mathrm{Holm}$ & HR \textit{own}:\textit{base} [95\% CI] & \textit{base} & \textit{own} & \textit{rand}\\
\midrule
FourRoom/LeWM & 40 & 0.53 & 0.70 & 0.50 & $+$0.17 [0.07, 0.28] & 0.02 & 1.99 [1.25, 3.43] & 0.20 & 0.38 & 0.05\\
Push-T/LeWM & 40 & 0.70 & 0.74 & 0.55 & $+$0.04 [$-$0.04, 0.11] & 0.95 & 1.10 [0.72, 1.62] & 0.33 & 0.29 & 0.25\\
Cube/LeWM & 40 & 0.23 & 0.19 & 0.24 & $-$0.04 [$-$0.14, 0.06] & 0.95 & 0.78 [0.37, 1.56] & 0.03 & 0.03 & 0.03\\
AntMaze/LeWM & 40 & 0.54 & 0.68 & 0.53 & $+$0.14 [0.00, 0.28] & 0.26 & 1.82 [1.16, 3.05] & 0.03 & 0.07 & 0.03\\
AntMaze/H-JEPA$^\dagger$ & 40 & 0.86 & 0.86 & 0.84 & 0.00 [$-$0.07, 0.07] & 1.00 & 1.05 [0.79, 1.48] & 0.62 & 0.65 & 0.60\\
\bottomrule
\end{tabular}
\end{table*}

\begin{table}[t]
\centering
\caption{\textbf{Trap rescue, pooled models} (stratified by trap; CMH stratified
by benchmark).}
\label{tab:rescue-pooled}
\setlength{\tabcolsep}{2pt}
\scriptsize
\begin{tabular}{l cc}
\toprule
 & LeWM cells (160 traps) & LeWM + AntMaze/H-JEPA (200 traps)\\
\midrule
\multicolumn{3}{l}{\textit{Stratified Cox HR (escape time)}}\\
\quad \textit{own} : \textit{base} & 1.45 [1.10, 1.88], $p$=0.002 & 1.32 [1.10, 1.64], $p$=0.005\\
\quad \textit{rand} : \textit{base} & 0.80 [0.60, 1.03], $p$=0.08 & 0.84 [0.68, 1.05], $p$=0.10\\
\quad \textit{own} : \textit{rand} & 1.80 [1.46, 2.38], $p$$<$10$^{-4}$ & 1.57 [1.28, 1.91], $p$$<$10$^{-4}$\\
\multicolumn{3}{l}{\textit{Cond.\ logit OR, escape@$H$}}\\
\quad \textit{own} : \textit{base} & 1.91 [1.22, 3.00], $p$=0.005 & 1.80 [1.17, 2.75], $p$=0.007\\
\quad \textit{rand} : \textit{base} & 0.70 [0.44, 1.09], $p$=0.11 & 0.69 [0.45, 1.05], $p$=0.09\\
\quad CMH \textit{own} : \textit{base} (cell strata) & 1.45 [1.03, 2.03], $p$=0.04 & 1.38 [1.01, 1.90], $p$=0.05\\
\multicolumn{3}{l}{\textit{Cond.\ logit OR, success from trap}}\\
\quad \textit{own} : \textit{base} & 2.80 [1.32, 5.92], $p$=0.007 & 1.96 [1.12, 3.42], $p$=0.02\\
\quad \textit{rand} : \textit{base} & 0.22 [0.09, 0.52], $p$$<$0.001 & 0.44 [0.25, 0.78], $p$=0.005\\
\bottomrule
\end{tabular}
\end{table}

\section{Additional analyses}
\label{app:extra}

\paragraph{Deeper hierarchies (HWM).}\label{app:hwm-res} We evaluate
HWM~\citep{zhang2026hwm} with two and three levels (HWM-L2, HWM-L3) using the
official checkpoints, planners and budgets (\cref{tab:hwm}). Unlike the
DINO-WM runs, these use per-cell configurations (\cref{app:hwm}). \EKR{}
improves success in all eight cells on the official tasks, significantly on
AntMaze for both models and on FourRoom with HWM-L2; the gains on Push-T and
Cube are positive but not significant. On held-out tasks with frozen
configurations all eight cells move in the same direction. Extra search does
not replace \EKR{} here either: on HWM-L2 AntMaze, $2\times$ and $4\times$
samples give 38.0\% and 35.3\% against 56.7\% for \EKR{} at $1.23\times$
compute, and on FourRoom the baseline needs $4$--$8\times$ samples (86.0\%
and 89.3\%) to reach the 89.3\% of \EKR{} at $2.06\times$.

\begin{table*}[t]
\centering
\caption{\textbf{HWM with two and three levels} (success \%, mean\,$\pm$\,std
over 3 seeds $\times$ 50 tasks; official checkpoints, planner and budgets).
Held-out tasks use a new task seed (10042) with frozen configurations.
$\Delta$: absolute gain of \EKR{} over the baseline. Exact McNemar test on 150
paired episodes: $^{*}p{<}0.05$, $^{**}p{<}0.01$, $^{***}p{<}0.001$.
Per-cell configurations are given in \cref{app:hwm}.}
\label{tab:hwm}
\setlength{\tabcolsep}{7pt}
\renewcommand{\arraystretch}{1.05}
\begin{tabular}{cllcccc}
\toprule
 & Backbone & Method & AntMaze & FourRoom & Push-T & Cube \\
\midrule
\multirow{6}{*}{\rotatebox[origin=c]{90}{\emph{Official tasks}}}
 & \multirow{3}{*}{HWM-L3}
 & Baseline    & \pmstd{28.7}{4.6} & \pmstd{97.3}{2.3} & \pmstd{12.7}{7.6} & \pmstd{58.0}{3.5} \\
 & & + \EKR{} (ours) & \textbf{\pmstd{44.7}{8.1}}$^{**}$ & \textbf{\pmstd{98.7}{1.2}} & \textbf{\pmstd{16.7}{1.2}} & \textbf{\pmstd{64.7}{6.1}} \\
 & & $\Delta$    & \up{16.0}{0} & \up{1.3}{0} & \up{4.0}{0} & \up{6.7}{0} \\
\cmidrule(l){2-7}
 & \multirow{3}{*}{HWM-L2}
 & Baseline    & \pmstd{34.0}{7.2} & \pmstd{70.7}{9.5} & \pmstd{49.3}{3.1} & \pmstd{47.3}{12.1} \\
 & & + \EKR{} (ours) & \textbf{\pmstd{56.7}{4.2}}$^{***}$ & \textbf{\pmstd{89.3}{4.2}}$^{***}$ & \textbf{\pmstd{56.0}{6.9}} & \textbf{\pmstd{49.3}{7.6}} \\
 & & $\Delta$    & \up{22.7}{0} & \up{18.7}{0} & \up{6.7}{0} & \up{2.0}{0} \\
\midrule
\multirow{6}{*}{\rotatebox[origin=c]{90}{\emph{Held-out tasks}}}
 & \multirow{3}{*}{HWM-L3}
 & Baseline    & \pmstd{35.3}{8.1} & \pmstd{94.0}{3.5} & \pmstd{8.0}{2.0} & \pmstd{52.7}{4.2} \\
 & & + \EKR{} (ours) & \textbf{\pmstd{47.3}{7.0}}$^{*}$ & \textbf{\pmstd{98.7}{2.3}} & \textbf{\pmstd{10.0}{4.0}} & \textbf{\pmstd{54.0}{6.0}} \\
 & & $\Delta$    & \up{12.0}{0} & \up{4.7}{0} & \up{2.0}{0} & \up{1.3}{0} \\
\cmidrule(l){2-7}
 & \multirow{3}{*}{HWM-L2}
 & Baseline    & \pmstd{34.0}{2.0} & \pmstd{70.0}{11.1} & \pmstd{44.7}{1.2} & \pmstd{48.0}{4.0} \\
 & & + \EKR{} (ours) & \textbf{\pmstd{59.3}{2.3}}$^{***}$ & \textbf{\pmstd{86.0}{3.5}}$^{***}$ & \textbf{\pmstd{49.3}{10.3}} & \textbf{\pmstd{51.3}{4.6}} \\
 & & $\Delta$    & \up{25.3}{0} & \up{16.0}{0} & \up{4.7}{0} & \up{3.3}{0} \\
\bottomrule
\end{tabular}
\end{table*}

\paragraph{Technical notes on edge cases.}\label{app:limits}
(i)~\emph{Repulsion near the goal.} On AntMaze/H-JEPA (seeds 42--43),
reached-then-pushed-away failures occur in 1 of 100 episodes with \EKR{} and
in none with the baseline.
(ii)~\emph{Offset in the base cost.} $\cRef$ assumes that the base cost has no
large candidate-independent offset, so that its magnitude reflects the
differences between candidates. For costs accumulated over many steps this may
not hold; an offset-invariant reference such as
$\cRef=\sg(\mathrm{mean}_i|\Cz^{(i)}-\min_{i'}\Cz^{(i')}|)$ can be used
instead.

\paragraph{Top-$k$ versus mean kernel contrast.} The spread of kernel values
across candidates (``contrast'') is larger for top-$k$ than for the
mean kernel on Cube/H-JEPA (0.037 vs.\ 0.025), FourRoom/H-JEPA (0.037 vs.\
0.015) and AntMaze/LeWM (0.044 vs.\ 0.036), and similar on AntMaze/H-JEPA
(0.013 vs.\ 0.010). Despite this, and consistent with the success rates in
\cref{tab:ablation}, the two
aggregations behave similarly in practice.

\paragraph{Bandwidth.} In endpoint mode $\sigma$ is $\tau_\sigma$ times
the median candidate-to-nearest-memory distance, with $\tau_\sigma{=}1$
(H-JEPA on AntMaze, FourRoom and Push-T, and the unified configuration except LeWM) or
$\tau_\sigma{=}1.5$ (Cube). The median candidate thus has a kernel value of
$\exp(-1/2)\approx0.61$ or $\exp(-1/(2\cdot1.5^2))\approx0.80$ against its
nearest entry: the kernel is neither saturated nor vanishing for typical
candidates. In observation mode (LeWM on AntMaze, FourRoom and Push-T),
$\sigma$ is twice the median nearest-neighbour distance between observed
latents.

\section{Latent-space mechanism, rescue curves, and recurrence analyses}
\label{app:latent}

This appendix complements \cref{sec:method,sec:behavior,app:rescue} with
direct visualisations and full-sample statistics in the world model's latent
space.

\subsection{How memory repulsion reshapes latent costs}
\label{app:latent-cost}

\begin{figure}[htbp]
\centering
\includegraphics{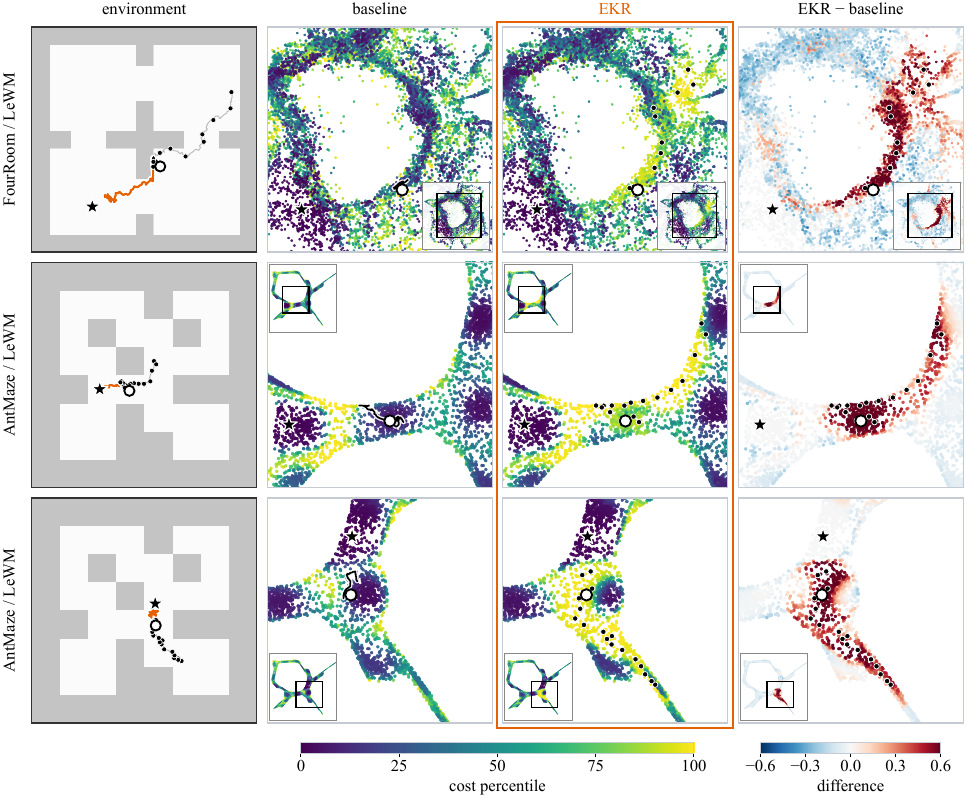}
\caption{\textbf{Single-replan cost shaping inside a trap.} Representative
baseline-fail / \EKR{}-succeed pairs (FourRoom/LeWM seed 42 task 0 at $t{=}110$;
AntMaze/LeWM seed 42 task 3 at $t{=}170$ and task 32 at $t{=}180$).
\emph{Environment:} path before the trap (thin grey) and inside it (thick
black), memory entries (black dots), agent (white circle), goal (star), and
subsequent \EKR{} path (orange). \emph{Baseline / \EKR{}:} full-latent cost
percentile on a per-seed 2-D Isomap layout (for display only), cropped to the
trap neighbourhood (full map in inset). \emph{\EKR{} $-$ baseline:} percentile
difference.}
\label{fig:app-costshaping}
\end{figure}

\begin{figure}[htbp]
\centering
\includegraphics{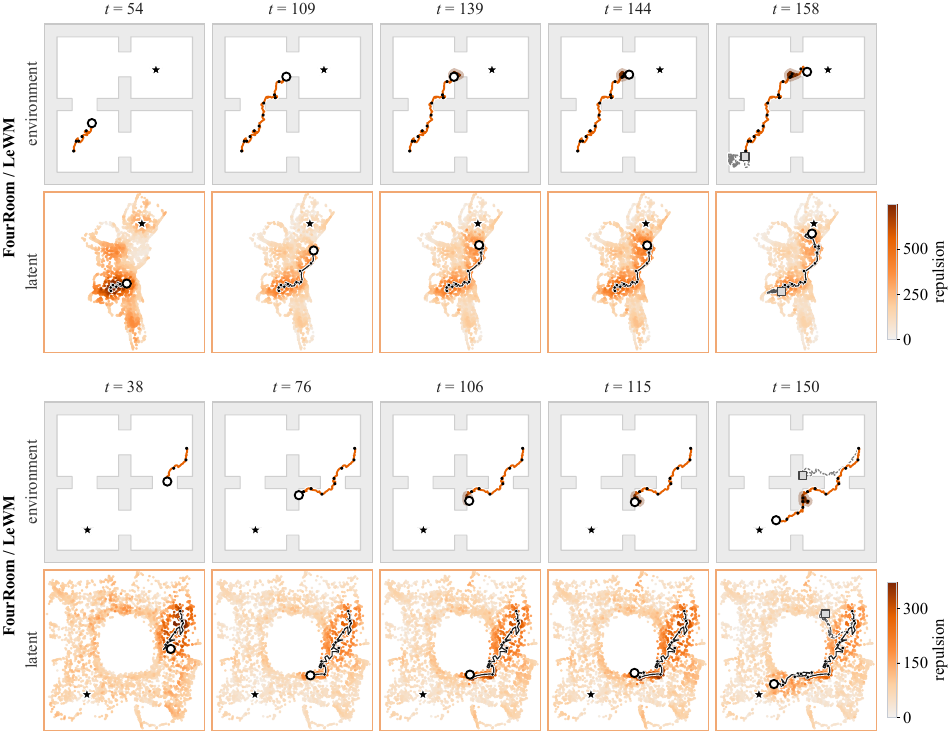}
\caption{\textbf{Memory repulsion over time as memory grows.} Two FourRoom/LeWM
episodes (top: seed 44, door 3 closed, $t\in\{54,109,139,144,158\}$ with
$M_t\in\{6,11,14,15,16\}$; bottom: seed 42, all doors open,
$t\in\{38,76,106,115,150\}$ with $M_t\in\{4,8,10,11,15\}$). Within each block,
top rows show the environment (trap segment shaded dark brown) and bottom rows
show full-latent repulsion above its per-panel minimum on a 2-D Isomap layout
(row-shared colour scale). The last column overlays the failing baseline
(dashed grey path, grey square) at the same step.}
\label{fig:app-timeline}
\end{figure}

\Cref{fig:app-costshaping} isolates a single replanning step inside a trap.
Under the goal cost alone, the trapped agent sits in a low-cost percentile
($10$th, $9$th, and $11$th in the three examples), so the baseline has no
incentive to leave. Adding memory repulsion lifts the agent's own state to the
$97$th, $95$th, and $96$th percentile while leaving the goal valley intact.
The shift is local: $67$--$69\%$ of latents within $0.5\sigma$ of memory rise
by at least $0.4$ in percentile, whereas points beyond $1\sigma$ on AntMaze
change by a median $|\Delta|$ of $0.05$. Even when AntMaze traps stall close
to the goal in $(x,y)$ (bottom row), those stalled states remain separated from
the goal in latent space.

\begin{figure}[htbp]
\centering
\includegraphics{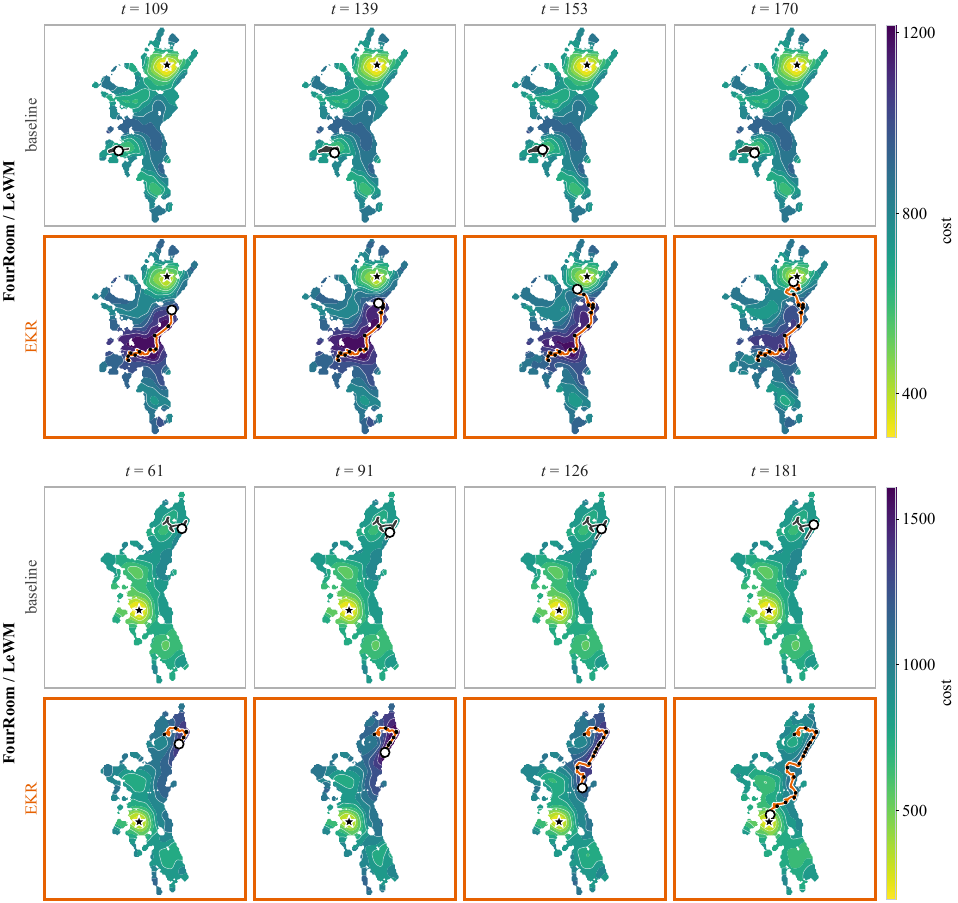}
\caption{\textbf{Planner cost terrain over time.} Two FourRoom/LeWM tasks (top:
seed 44, door 3 closed, $t\in\{109,139,153,170\}$; bottom: seed 42, door 3
closed, $t\in\{61,91,126,181\}$). Grey frames: static baseline goal cost $\Cz$;
orange frames: \EKR{} cost $\Ctot$ above its per-panel minimum, computed in the
full latent space and interpolated on a 2-D Isomap layout (white: no data;
block-shared colour scale).}
\label{fig:app-costterrain}
\end{figure}

\Cref{fig:app-timeline,fig:app-costterrain} track this shaping across full
episodes. Whether the trap is a corridor dead end (\cref{fig:app-timeline}, top)
or a latent loop (bottom), repulsion builds along the visited trap path: at
escape, mean repulsion on the trap trajectory is $2.8\times$ and $4.2\times$
that of the remaining seed states, steering \EKR{} to the goal while the
paired baseline stays stalled. In \cref{fig:app-costterrain}, \EKR{} raises the
trap trajectory by roughly half the cost range on average ($0.49$ and $0.51$),
making the route to the goal the lowest-cost direction. Notably, in the full
latent space the baseline's observed trap states are not always a deep local
minimum of $\Cz$ (nearest $2\%$ versus $5$--$15\%$ outer ring: $+0.12\text{ SD}$
in the top example, $-0.06\text{ SD}$ in the bottom); stalling arises from the
planner's rolled-out trajectories repeatedly favouring the same basin, which
history-dependent repulsion breaks regardless.

\subsection{Trap rescue and time-to-solve dynamics}
\label{app:latent-curves}

\begin{figure}[htbp]
\centering
\includegraphics{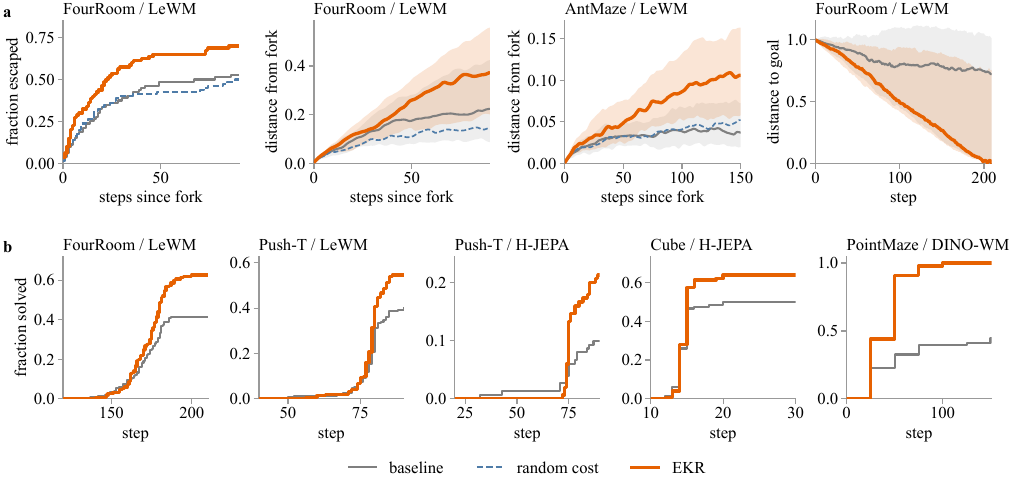}
\caption{\textbf{Trap-rescue and time-to-solve curves.} (a) Paired rescue from
$40$ deep traps $\times$ $2$ planner seeds ($80$ runs per branch): cumulative
escape fraction (FourRoom/LeWM), distance moved from the fork
(FourRoom/LeWM and AntMaze/LeWM; median and interquartile band; random-cost
median dashed), and normalised goal distance in the main FourRoom/LeWM runs.
(b) Fraction of tasks solved by step $t$ across the five settings where \EKR{}
significantly improves final success ($150$ paired episodes each; $x$-axes
start near the first success and $y$-axes end at $1.15\times$ the larger final
rate).}
\label{fig:app-rescue-curves}
\end{figure}

On the paired trap forks (\cref{fig:app-rescue-curves}a), \EKR{} raises
FourRoom/LeWM escape within $H{=}90$ from $0.525$ (baseline) and $0.500$
(random cost) to $0.700$ (exact McNemar $p{=}0.0005$), and increases median
distance from the fork at $H$ on both FourRoom ($0.224$ to $0.375$, Wilcoxon
$p{=}0.005$) and AntMaze ($0.036$ to $0.105$, $p{=}10^{-8}$). Across full
FourRoom/LeWM episodes, median normalised goal distance drops from $0.821$ to
$0.479$ ($p{=}7\times 10^{-6}$). Random cost fails to reproduce these gains,
confirming that experience-anchored repulsion—not cost perturbation—drives
escape. In the main evaluation (\cref{fig:app-rescue-curves}b), \EKR{} pulls
ahead once episodes reach the post-trap regime (e.g.\ PointMaze/DINO-WM
$0.447\to 1.000$, $p{=}2\times 10^{-25}$; FourRoom/LeWM $0.413\to 0.627$,
$p{=}6\times 10^{-6}$).

\subsection{Latent-space recurrence and visit topology}
\label{app:latent-recurrence}

\begin{figure}[htbp]
\centering
\includegraphics{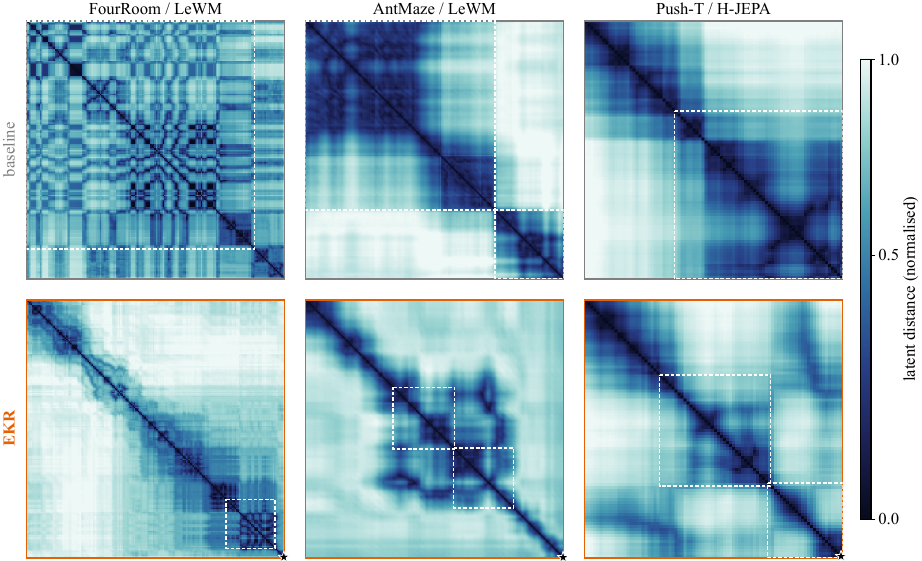}
\caption{\textbf{Self-distance matrices in the full latent space.} Pairwise
distances $\norm{z_i-z_j}_2$ normalised by the pair's $97$th percentile within
the same pre-success window (ending at \EKR{}'s goal step; same seed and task
per column: FourRoom/LeWM s42 t33, AntMaze/LeWM s42 t40, Push-T/H-JEPA s44
t47). Dashed boxes mark detected traps; stars mark goal arrival.}
\label{fig:app-selfdistance}
\end{figure}

\begin{table}[htbp]
\centering
\caption{\textbf{Latent-space recurrence over all paired episodes} (3 seeds
$\times$ 50 tasks; no selection). Both arms are truncated to the same
pre-success window; $\varepsilon$ is the seed-wise $10$th percentile of
within-episode full-latent distances. Values are medians over pairs; $p$:
paired Wilcoxon signed-rank test on the recurrence rate (fraction of pairs
with $|i-j|{>}10$ and distance ${<}\varepsilon$). Mean revisits: average
earlier steps ($s\le t{-}10$) within $\varepsilon$ of $z_t$.
$\varepsilon$-cover: greedy number of $\varepsilon$-balls covering the
trajectory.}
\label{tab:recurrence}
\setlength{\tabcolsep}{7pt}
\begin{tabular}{lccccc}
\toprule
Cell & Recurrence rate (base $\rightarrow$ \EKR{}) & \EKR{} lower & Wilcoxon $p$ & Mean revisits & $\varepsilon$-cover \\
\midrule
FourRoom/LeWM & 9.2\% $\rightarrow$ \textbf{4.3\%} & 69\% & $7.3\times10^{-9}$ & 8.6 $\rightarrow$ \textbf{4.0} & 16.5 $\rightarrow$ \textbf{23.5} \\
AntMaze/LeWM & 7.7\% $\rightarrow$ \textbf{4.2\%} & 63\% & $8.9\times10^{-4}$ & 11.6 $\rightarrow$ \textbf{6.4} & 22 $\rightarrow$ \textbf{30} \\
Push-T/H-JEPA & 9.7\% $\rightarrow$ \textbf{4.6\%} & 61\% & $0.003$ & 3.7 $\rightarrow$ \textbf{1.8} & 11 $\rightarrow$ \textbf{14} \\
\bottomrule
\end{tabular}
\end{table}

\begin{figure}[htbp]
\centering
\hspace*{10pt}\includegraphics{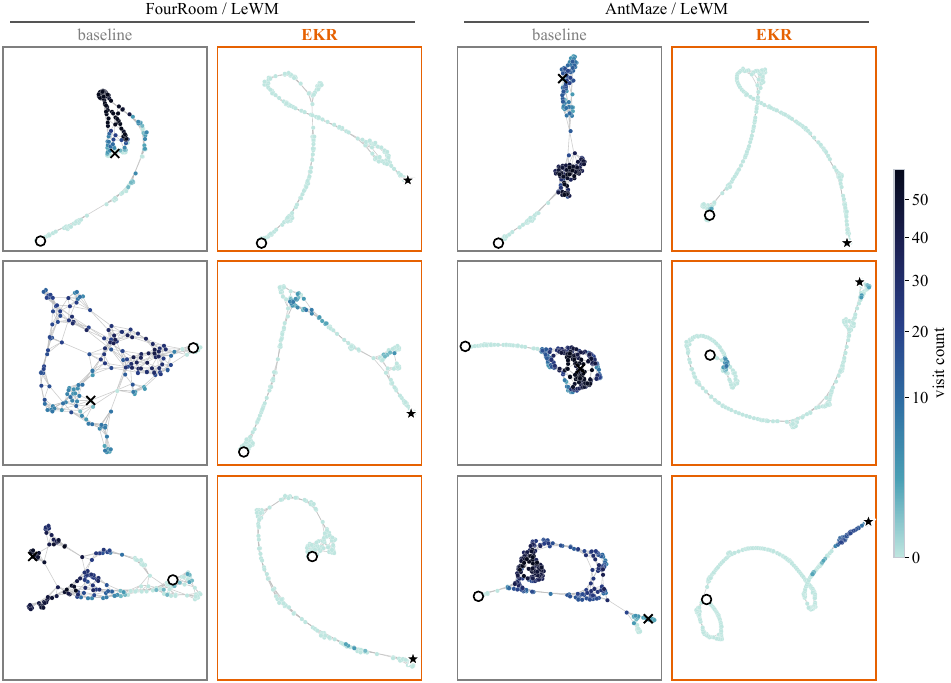}
\caption{\textbf{Latent visit graphs.} Force-directed layouts of $6$-nearest-neighbour
graphs built in the full latent space (each node is one step; no latent
projection). Each row-pair compares the failing baseline and successful \EKR{}
on the same task and window (equal node counts). Colour encodes visit count
(capped at the $97$th percentile); markers show start ($\circ$), failed end
($\times$), and goal ($\star$).}
\label{fig:app-visitgraph}
\end{figure}

\begin{figure}[htbp]
\centering
\includegraphics{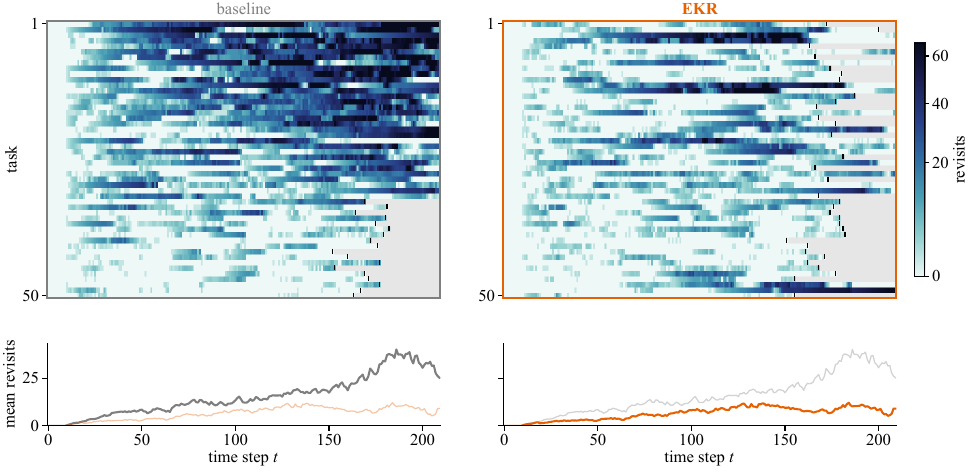}
\caption{\textbf{Step-by-step revisits across all 50 tasks of FourRoom/LeWM
(seed 44).} Rows are tasks aligned between panels and sorted by baseline mean
revisits. Colour counts earlier steps ($s\le t{-}10$) within $\varepsilon$ of
$z_t$; black ticks mark goal arrival, and light grey marks completed episodes.
Bottom curves show mean revisits over running tasks (faint curve: the other
method).}
\label{fig:app-revisitheatmap}
\end{figure}

Finally, \cref{fig:app-selfdistance,tab:recurrence,fig:app-visitgraph,fig:app-revisitheatmap}
quantify latent revisits without relying on 2-D projections. In self-distance
matrices (\cref{fig:app-selfdistance}), baselines form dark off-diagonal
blocks as the agent cycles in place (recurrence $9.3\%$, $10.5\%$, and
$2.6\%$), whereas \EKR{} stays near the diagonal ($1.9\%$, $1.5\%$, and
$0.0\%$) and exits traps rapidly ($35$ steps on FourRoom; $68$ and $66$ on
AntMaze). Across all $150$ uncherry-picked pairs per cell
(\cref{tab:recurrence}), \EKR{} roughly halves the median recurrence rate and
mean revisits while expanding $\varepsilon$-coverage ($p\le 0.003$ throughout).

Topologically (\cref{fig:app-visitgraph}), baseline trajectories collapse into
dense, high-revisit knots, whereas \EKR{} unfolds into near-linear chains from
start to goal with $1.7$--$3.4\times$ larger hop diameter at identical node
counts. Across all $50$ tasks of FourRoom/LeWM seed 44 without episode
selection (\cref{fig:app-revisitheatmap}), \EKR{} cuts mean revisits from
$14.4$ to $6.1$ and raises solved tasks from $18$ to $33$, though a few
tasks still retain dark bands where traps persist.

\clearpage
\section{Extended sensitivity and hyperparameter robustness}
\label{app:sensitivity}

Unless noted otherwise (when varying the corresponding hyperparameter), all
experiments in \cref{app:sensitivity,app:diagnostics,app:metaworld} use the
unified default configuration ($K{=}3$, $\lambda{=}1.5$, $\tau_\sigma{=}1$,
memory capacity $M{=}128$, novelty gate $\epsilon_{\text{nov}}{=}0.1$) with no
per-task tuning, and report $\text{mean}\pm\text{std}$ (\%) over $3$ planning
seeds $\times$ $50$ tasks ($150$ episodes per cell) with two-sided exact
McNemar $p$-values on pooled discordant pairs.

\subsection{Two-sided sensitivity window in repulsion weight \texorpdfstring{$\lambda$}{lambda}}
\label{app:lambda_sweep}

\begin{figure}[htbp]
\centering
\includegraphics{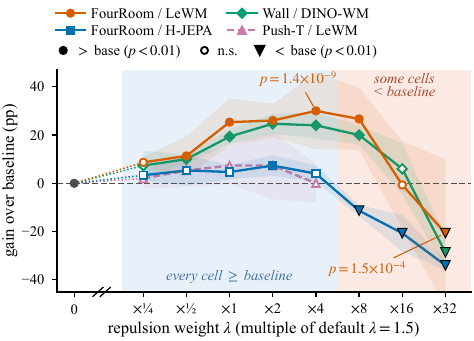}
\caption{\textbf{Two-sided $\lambda$ window.} Gain of \EKR{} over baseline (pp) as $\lambda$ scales from $\frac{1}{4}\times$ to $32\times$ the default $\lambda{=}1.5$ ($\log_2$ axis; $\lambda{=}0$ is baseline). Shaded bands: $\pm 1$\,s.d.\ of per-seed paired gains. Filled markers: $p{<}0.01$ above baseline; open: n.s.; downward triangles: $p{<}0.01$ below baseline.}
\label{fig:lambda_window}
\end{figure}

\Cref{tab:lambda_sweep,fig:lambda_window} scale $\lambda$ from
$\times 0.25$ ($\lambda{=}0.375$) to $\times 32$ ($\lambda{=}48$) across three
world-model backbones (LeWM, H-JEPA, DINO-WM) and two planner families
(gradient descent and CEM). Across FourRoom/LeWM, FourRoom/H-JEPA,
Wall-long/DINO-WM, and TwoRoom-long/LeWM, performance follows the two-sided
window predicted in \cref{app:theory-random}: repulsion weights spanning
$16\times$--$32\times$ in range ($\times 0.25$ to $\times 8$ around
$\lambda{=}1.5$) consistently improve over or match the baseline, whereas
extreme weights ($\times 16$--$\times 32$) overpower the goal cost and drop
below the baseline. On Push-T/LeWM (swept to $\times 4$), \EKR{} stays at or
above the $40.0\%$ baseline ($40.0\%$--$47.3\%$).

\begin{table}[htbp]
\centering
\small
\caption{\textbf{Sensitivity to the repulsion weight $\lambda$} (3 seeds $\times$ 50 tasks per cell). \textbf{Bold}: significantly above baseline (exact McNemar, $p{<}0.01$, except TwoRoom-long where $p{=}0.011$ at $\times 1$ and $p{=}0.028$ at $\times 4$); \textit{italic}$^\downarrow$: significantly below baseline ($p{<}0.01$).}
\label{tab:lambda_sweep}
\resizebox{\linewidth}{!}{%
\begin{tabular}{l c c c c c c c c c}
\toprule
Cell & Base & $\times0.25$ & $\times0.5$ & $\times1$ & $\times2$ & $\times4$ & $\times8$ & $\times16$ & $\times32$ \\
 & & \scriptsize $\lambda{=}0.375$ & \scriptsize $\lambda{=}0.75$ & \scriptsize $\lambda{=}1.5$ & \scriptsize $\lambda{=}3$ & \scriptsize $\lambda{=}6$ & \scriptsize $\lambda{=}12$ & \scriptsize $\lambda{=}24$ & \scriptsize $\lambda{=}48$ \\
\midrule
FourRoom/LeWM & 41.3 & 50.0$\pm$12.5 & \textbf{52.7$\pm$8.3} & \textbf{66.7$\pm$5.0} & \textbf{67.3$\pm$8.1} & \textbf{71.3$\pm$1.2} & \textbf{68.0$\pm$5.3} & 40.7$\pm$7.0 & \textit{20.7$\pm$16.0}$^{\downarrow}$ \\
FourRoom/H-JEPA & 92.7 & 96.0$\pm$2.0 & 98.0$\pm$2.0 & 97.3$\pm$2.3 & \textbf{100.0$\pm$0.0} & 96.7$\pm$3.1 & \textit{81.3$\pm$4.2}$^{\downarrow}$ & \textit{72.0$\pm$4.0}$^{\downarrow}$ & \textit{58.7$\pm$4.2}$^{\downarrow}$ \\
Wall-long/DINO-WM & 75.3 & \textbf{82.7$\pm$3.1} & \textbf{85.3$\pm$3.1} & \textbf{94.7$\pm$2.3} & \textbf{100.0$\pm$0.0} & \textbf{99.3$\pm$1.2} & \textbf{95.3$\pm$1.2} & 81.3$\pm$7.0 & \textit{46.7$\pm$2.3}$^{\downarrow}$ \\
TwoRoom-long/LeWM & 51.3 & 50.7$\pm$2.3 & 56.7$\pm$3.1 & \textbf{61.3$\pm$4.2} & 58.7$\pm$3.1 & \textbf{61.3$\pm$5.8} & 54.7$\pm$1.2 & 48.7$\pm$5.0 & 43.3$\pm$4.6 \\
Push-T/LeWM & 40.0 & 42.0$\pm$2.0 & 45.3$\pm$4.6 & 47.3$\pm$10.1 & 47.3$\pm$11.0 & 40.0$\pm$5.3 & -- & -- & -- \\
\bottomrule
\end{tabular}%
}
\end{table}

\subsection{Robustness to neighbour count \texorpdfstring{$K$}{K} and bandwidth \texorpdfstring{$\tau_\sigma$}{tau\_sigma}}
\label{app:k_sigma}

\begin{figure}[htbp]
\centering
\includegraphics{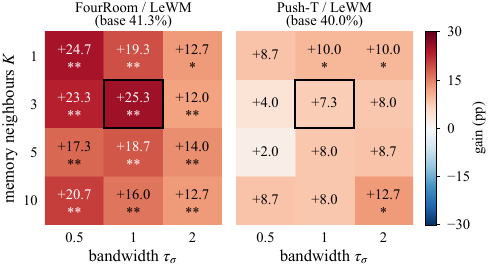}
\caption{\textbf{Robustness to $K$ and $\tau_\sigma$.} Gain over baseline (pp) on FourRoom/LeWM (left, base $41.3\%$) and Push-T/LeWM (right, base $40.0\%$). $^{**}\,p{<}0.01$, $^{*}\,p{<}0.05$; black box marks the unified configuration ($K{=}3, \tau_\sigma{=}1$).}
\label{fig:k_sigma_heatmap}
\end{figure}

\Cref{tab:k_sigma,fig:k_sigma_heatmap} evaluate the $4\times 3$ grid of
$K\in\{1,3,5,10\}$ and $\tau_\sigma\in\{0.5,1,2\}$ across four model--task
cells. All $48/48$ grid points match or exceed their respective baselines:
all $12$ points on FourRoom/LeWM are significantly above baseline
($p\in[2.5\times 10^{-8}, 0.011]$), all $12$ on Wall-long/DINO-WM exceed
$75.3\%$ ($81.3\%$--$95.3\%$, with $8$ points at $p\le 2\times 10^{-5}$), all
$12$ on FourRoom/H-JEPA reach $96.0\%$--$99.3\%$ (vs.\ $92.7\%$), and all $12$
on Push-T/LeWM reach $42.0\%$--$52.7\%$ (vs.\ $40.0\%$, $3$ points $p{<}0.05$).

\begin{table}[htbp]
\centering
\small
\caption{\textbf{Robustness to $K$ and $\tau_\sigma$} ($\lambda{=}1.5$ fixed, 3 seeds $\times$ 50 tasks). \textbf{Bold}: $p{<}0.05$ over baseline (pooled exact McNemar, uncorrected). $^{\dagger}$\,FourRoom/H-JEPA at $(K{=}5,\tau_\sigma{=}2)$ has $n{=}148$ (seed 44 completed 48/50 tasks); all other 47 points have $n{=}150$.}
\label{tab:k_sigma}
\begin{tabular}{l c ccc}
\toprule
Cell & $K$ & $\tau_\sigma{=}0.5$ & $\tau_\sigma{=}1$ & $\tau_\sigma{=}2$ \\
\midrule
FourRoom/LeWM (base 41.3) & 1 & \textbf{66.0$\pm$9.2} & \textbf{60.7$\pm$6.4} & \textbf{54.0$\pm$10.6} \\
 & 3 & \textbf{64.7$\pm$8.1} & \textbf{66.7$\pm$5.0} & \textbf{53.3$\pm$8.1} \\
 & 5 & \textbf{58.7$\pm$8.1} & \textbf{60.0$\pm$10.6} & \textbf{55.3$\pm$9.5} \\
 & 10 & \textbf{62.0$\pm$6.0} & \textbf{57.3$\pm$7.0} & \textbf{54.0$\pm$6.9} \\
\midrule
Push-T/LeWM (base 40.0) & 1 & 48.7$\pm$10.3 & \textbf{50.0$\pm$5.3} & \textbf{50.0$\pm$2.0} \\
 & 3 & 44.0$\pm$3.5 & 47.3$\pm$10.1 & 48.0$\pm$6.9 \\
 & 5 & 42.0$\pm$5.3 & 48.0$\pm$5.3 & 48.7$\pm$6.1 \\
 & 10 & 48.7$\pm$8.1 & 48.0$\pm$5.3 & \textbf{52.7$\pm$6.4} \\
\midrule
Wall-long/DINO-WM (base 75.3) & 1 & \textbf{83.3$\pm$3.1} & \textbf{95.3$\pm$3.1} & \textbf{90.0$\pm$3.5} \\
 & 3 & 81.3$\pm$1.2 & \textbf{94.7$\pm$2.3} & \textbf{89.3$\pm$4.2} \\
 & 5 & 82.0$\pm$2.0 & \textbf{92.0$\pm$0.0} & \textbf{90.0$\pm$2.0} \\
 & 10 & 81.3$\pm$3.1 & \textbf{91.3$\pm$1.2} & \textbf{90.0$\pm$2.0} \\
\midrule
FourRoom/H-JEPA (base 92.7) & 1 & \textbf{99.3$\pm$1.2} & 96.0$\pm$5.3 & 98.0$\pm$2.0 \\
 & 3 & \textbf{98.7$\pm$2.3} & 97.3$\pm$2.3 & \textbf{99.3$\pm$1.2} \\
 & 5 & \textbf{98.0$\pm$2.0} & 98.0$\pm$2.0 & 98.0$\pm$2.8$^{\dagger}$ \\
 & 10 & 96.7$\pm$2.3 & \textbf{98.7$\pm$1.2} & \textbf{99.3$\pm$1.2} \\
\bottomrule
\end{tabular}
\end{table}

\subsection{Robustness to memory capacity \texorpdfstring{$M$}{M} and novelty gate \texorpdfstring{$\epsilon_{\text{nov}}$}{epsilon\_nov}}
\label{app:mem_tol}

\Cref{tab:mem_tol} ablates capacity $M$ and novelty gate $\epsilon_{\text{nov}}$
on FourRoom/LeWM. Storing even $M{=}4$ visited latents raises success from
$41.3\%$ to $62.7\%$ ($p{=}1.8\times 10^{-6}$), and all tested values remain
highly significant ($p\le 1.5\times 10^{-5}$). Because FourRoom/LeWM adds at
most $22$ entries per episode (max $=$ median $= 22$ across all $50$ tasks on
seed 42), any $M\ge 22$ (including $M{=}128$ and the codebase default $256$)
is structurally identical, verified bitwise on seed 42. A single-seed pilot on
Wall-long/DINO-WM (seed 0, base $72.0\%$, max $6$ additions per episode) shows
the same insensitivity: $M\in\{2,4,8\}$ all reach $92.0\%$, and
$\epsilon_{\text{nov}}\in\{0.05,0.2,0.4\}$ reach $92.0\%/92.0\%/90.0\%$.

\begin{table}[htbp]
\centering
\small
\caption{\textbf{Robustness to memory capacity $M$ and novelty gate $\epsilon_{\text{nov}}$ on FourRoom/LeWM} (3 seeds $\times$ 50 tasks). $^{\star}$\,Because at most 22 entries are added per episode, $M{\ge}32$ is structurally identical to $M{=}16$ and the unified default ($M{=}128$).}
\label{tab:mem_tol}
\begin{tabular}{l l c c c}
\toprule
Hyperparameter & Value & Success (\%) & Disc.\ (\EKR{}:base) & $p$ \\
\midrule
Baseline ($\lambda{=}0$) & -- & 41.3$\pm$14.6 & -- & -- \\
\midrule
Memory capacity $M$ & $4$ & \textbf{62.7$\pm$5.0} & 39\,:\,7 & $1.8\times 10^{-6}$ \\
 & $8$ & \textbf{65.3$\pm$6.1} & 42\,:\,6 & $1.0\times 10^{-7}$ \\
 & $16$ & \textbf{66.7$\pm$5.0} & 44\,:\,6 & $3.2\times 10^{-8}$ \\
 & $32\,/\,64\,/\,128^{\star}$ (default) & \textbf{66.7$\pm$5.0} & 44\,:\,6 & $3.2\times 10^{-8}$ \\
\midrule
Novelty gate $\epsilon_{\text{nov}}$ & $0.05$ & \textbf{66.7$\pm$5.0} & 44\,:\,6 & $3.2\times 10^{-8}$ \\
 & $0.10$ (default) & \textbf{66.7$\pm$5.0} & 44\,:\,6 & $3.2\times 10^{-8}$ \\
 & $0.20$ & \textbf{66.7$\pm$4.6} & 45\,:\,7 & $7.0\times 10^{-8}$ \\
 & $0.40$ & \textbf{60.7$\pm$11.4} & 37\,:\,8 & $1.5\times 10^{-5}$ \\
\bottomrule
\end{tabular}
\end{table}

\section{Diagnostic mechanism analysis}
\label{app:diagnostics}

To examine why \EKR{} helps and which failure modes it removes, we analyse
per-step offline evaluation logs (AntMaze: seeds 42--43, $n{=}100$; all other
cells: seeds 42--44, $n{=}150$--$300$).

\subsection{Latent goal cost plateaus far from the goal}
\label{app:latent_plateau}

\begin{figure}[H]
\centering
\includegraphics{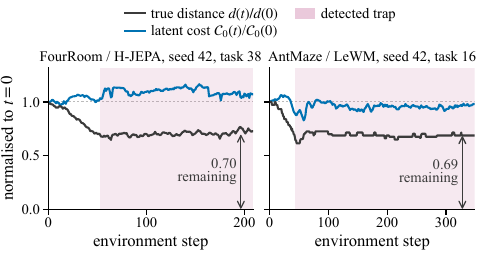}
\caption{\textbf{The latent cost plateaus far from the goal.} Two real baseline episodes: true geodesic distance $d(t)$ (black) and latent goal cost $\Cz(t)$ (blue), normalized by $t{=}0$ (shading: detected trap). Left: FourRoom/H-JEPA (seed 42, task 38); right: AntMaze/LeWM (seed 42, task 16). \EKR{} solves both under the same seed (steps 129 and 162).}
\label{fig:latent_plateau}
\end{figure}

\Cref{tab:m2_latent_cost,fig:latent_plateau} show that across baseline
episodes, the Spearman correlation $\rho_s(\Cz, d)$ between latent goal cost
and true task distance is moderate ($0.46$--$0.68$). During detected traps,
most of the true distance remains ($\tilde{d}_{\text{trap}}=0.78$--$1.00$),
while $\Cz$ plateaus near its initial value
($\tilde{\mathcal{C}}_{0,\text{trap}}/\mathcal{C}_0(0)=0.95$--$0.97$) with
low within-trap variation ($\text{CV}_{\text{trap}}=0.06$ on AntMaze,
FourRoom, and Push-T; $0.19$ on Cube).

\begin{table}[htbp]
\centering
\small
\caption{\textbf{Latent goal cost $\Cz$ versus true task distance $d$ in baseline episodes.} During trap events, true distance remains large while $\Cz$ plateaus near $\Cz(0)$.}
\label{tab:m2_latent_cost}
\begin{tabular}{lrrrrrr}
\toprule
Env. & episodes & $\rho_s(c_0,d)$ & trap events & $\tilde d_{\text{trap}}$ & $\tilde c_{0,\text{trap}}/c_0(0)$ & CV$_{\text{trap}}$ \\
\midrule
AntMaze & 200 & 0.68 & 276 & 0.82 & 0.96 & 0.06 \\
FourRoom & 300 & 0.59 & 275 & 0.78 & 0.95 & 0.06 \\
Push-T & 300 & 0.46 & 359 & 0.96 & 0.97 & 0.06 \\
Cube & 300 & 0.49 & 248 & 1.00 & 0.97 & 0.19 \\
\bottomrule
\end{tabular}
\end{table}

\subsection{Gain by goal distance and failure-type breakdown}
\label{app:failure_types}

\begin{figure}[H]
\centering
\includegraphics{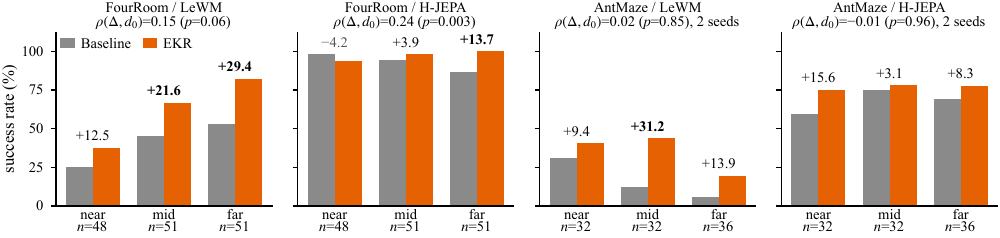}
\caption{\textbf{Gain versus goal distance (navigation cells).} Baseline (grey) and \EKR{} (orange) success rate across near / mid / far tertiles of initial geodesic distance $d_0$. Bold gain labels: 95\% paired bootstrap CI excludes 0.}
\label{fig:gain_distance}
\end{figure}

\Cref{tab:s14_gain_distance,fig:gain_distance} bin navigation episodes into
tertiles of initial geodesic distance $d_0$. On FourRoom, \EKR{}'s gain grows
monotonically with distance for both backbones (LeWM:
$+12.5\to +21.6\to +29.4$\,pp; H-JEPA: $-4.2\to +3.9\to +13.7$\,pp,
$\rho_s{=}0.24, p{=}0.003$); on AntMaze, \EKR{} improves every tertile without
a monotonic trend. \Cref{tab:s5_failure_types} decomposes paired outcomes: on LeWM, \EKR{} halves ``no-start'' traps (FourRoom
$62\to 31$, AntMaze $59\to 34$); on H-JEPA, it mainly removes mid-episode traps
(FourRoom $7\to 1$, AntMaze $19\to 14$). Across all four cells, ``pushed away''
failures remain at $0$--$1$ episodes. Taken together with \cref{app:latent_plateau}, these diagnostics show a consistent pattern: when $\Cz$ stalls near its initial value far from the goal, memory repulsion breaks early no-start traps on flat models (LeWM) and mid-episode traps on hierarchical models (H-JEPA), yielding larger gains on longer-horizon navigation tasks without destabilising successful runs.

\begin{table}[H]
\centering
\small
\caption{\textbf{Success rate by start--goal geodesic distance tertile} (base$\to$\EKR{} \%, gain in pp; AntMaze: 2 seeds; FourRoom: 3 seeds).}
\label{tab:s14_gain_distance}
\begin{tabular}{llrcccr}
\toprule
Cell & $n$ & seeds & near & mid & far & $\rho_s(\Delta,d_0)$ \\
\midrule
FourRoom/LeWM & 150 & 42,43,44 & 25$\to$38 (+12.5) & 45$\to$67 (+21.6) & 53$\to$82 (+29.4) & 0.15 ($p{=}0.062$) \\
FourRoom/H-JEPA & 150 & 42,43,44 & 98$\to$94 ($-$4.2) & 94$\to$98 (+3.9) & 86$\to$100 (+13.7) & 0.24 ($p{=}0.003$) \\
AntMaze/LeWM & 100 & 42,43 & 31$\to$41 (+9.4) & 12$\to$44 (+31.2) & 6$\to$19 (+13.9) & 0.02 ($p{=}0.847$) \\
AntMaze/H-JEPA & 100 & 42,43 & 59$\to$75 (+15.6) & 75$\to$78 (+3.1) & 69$\to$78 (+8.3) & $-$0.01 ($p{=}0.960$) \\
\bottomrule
\end{tabular}
\end{table}

\begin{table}[H]
\centering
\small
\caption{\textbf{Paired episode outcome counts} (baseline$\to$\EKR{}) on navigation cells.}
\label{tab:s5_failure_types}
\begin{tabular}{lrccccc}
\toprule
Cell & $n$ & success & progressing & pushed away & mid-trap & no-start \\
\midrule
FourRoom/LeWM & 150 & 62$\to$94 & 6$\to$15 & 0$\to$0 & 20$\to$10 & 62$\to$31 \\
FourRoom/H-JEPA & 150 & 139$\to$146 & 1$\to$2 & 0$\to$0 & 7$\to$1 & 3$\to$1 \\
AntMaze/LeWM & 100 & 16$\to$34 & 2$\to$14 & 1$\to$1 & 22$\to$17 & 59$\to$34 \\
AntMaze/H-JEPA & 100 & 68$\to$77 & 11$\to$6 & 0$\to$1 & 19$\to$14 & 2$\to$2 \\
\bottomrule
\end{tabular}
\end{table}

\section{Metaworld \texttt{mw-reach-wall} and exclusive-GPU timing}
\label{app:metaworld}

\begin{figure}[H]
\centering
\includegraphics{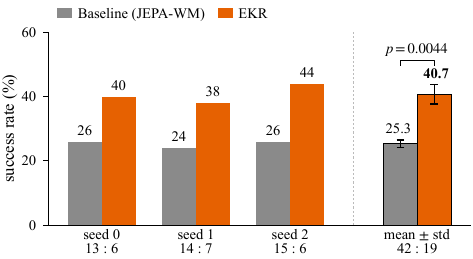}
\caption{\textbf{JEPA-WM on Metaworld \texttt{mw-reach-wall}.} Per-seed (0--2) and pooled success rates under the official protocol ($25.3\pm 1.2\% \to 40.7\pm 3.1\%$, pooled discordant pairs $42:19$, $p{=}0.0044$).}
\label{fig:metaworld}
\end{figure}

\paragraph{Generalization to Metaworld \texttt{mw-reach-wall}.}
We evaluate the unified \EKR{} configuration without tuning on the official
\texttt{jepa-wms} checkpoint and protocol for Metaworld \texttt{mw-reach-wall}
(first $50$ official episodes, $100$ steps, final-step success, $3$ planning
seeds). As shown in \cref{tab:metaworld_jepa,fig:metaworld}, \EKR{} raises
success from $25.3\pm 1.2\%$ to $40.7\pm 3.1\%$ (pooled discordant pairs
$42:19$, exact McNemar $p{=}0.0044$), leading on every seed ($+14$, $+14$,
$+18$\,pp). Measuring whether the agent reaches the goal at any step shows an
increase from $35.3\%$ to $61.3\%$ ($p{=}4.6\times 10^{-7}$), consistent with
escaping local traps.

\begin{table}[H]
\centering
\small
\caption{\textbf{Metaworld \texttt{mw-reach-wall} with the official JEPA-WM checkpoint and protocol} (3 seeds $\times$ 50 official episodes).}
\label{tab:metaworld_jepa}
\begin{tabular}{llcccc}
\toprule
Model & Task & Base & +\EKR{} & Disc.\ (\EKR{}:base) & $p$ \\
\midrule
JEPA-WM & mw-reach-wall & $25.3\pm 1.2$ & $\mathbf{40.7\pm 3.1}$ & 42\,:\,19 & $0.0044$ \\
\bottomrule
\end{tabular}
\end{table}

\paragraph{Per-replan wall-clock overhead on an exclusive GPU.}
On an exclusive A100 80\,GB GPU (seed 42, $6$ alternating episodes per arm
after $2$ warmup plans), \cref{tab:timing_clean} shows that the repulsion term
adds only $25.3$--$25.9$\,ms per replan ($<1\%$ of plan time), leaves
per-iteration solver time unchanged within noise ($+4.5\%$ $[-0.7, +6.8]$ on
FourRoom; $-1.0\%$ $[-7.1, +6.7]$ on Push-T), and uses identical peak GPU
memory ($1.89$\,GB on FourRoom; $3.36$\,GB on Push-T).

\begin{table}[H]
\centering
\small
\caption{\textbf{Per-replanning-step wall-clock on an exclusive A100 GPU} (median [IQR]). Repulsion adds ${\sim}25$\,ms per replan ($<$1\% of plan time) with unchanged per-iteration solver cost; fewer solver iterations yield lower total plan time.}
\label{tab:timing_clean}
\begin{tabular}{llcccc}
\toprule
Cell & Arm & Plan time (s) & Iters/plan & ms/iter & Repulsion (ms) \\
\midrule
FourRoom/LeWM & base & 8.14 [4.04, 14.26] & 50.5 & 159 & -- \\
              & \EKR{}  & 5.99 [3.04, 8.85]  & 32.0 & 166 & 25.3 \\
              & \multicolumn{5}{l}{\footnotesize overhead per iter: $+4.5\%$ [$-0.7$, $+6.8$]; repulsion share 0.45\%} \\
\midrule
Push-T/LeWM   & base & 5.15 [2.94, 9.29]  & 58.0 & 105 & -- \\
              & \EKR{}  & 4.09 [1.66, 7.21]  & 34.5 & 104 & 25.9 \\
              & \multicolumn{5}{l}{\footnotesize overhead per iter: $-1.0\%$ [$-7.1$, $+6.7$]; repulsion share 0.70\%} \\
\bottomrule
\end{tabular}
\end{table}

\end{document}